\documentclass{article}

\PassOptionsToPackage{numbers, square}{natbib}

\usepackage[main, final]{neurips_2026}

\usepackage[utf8]{inputenc} % allow utf-8 input
\usepackage[T1]{fontenc}    % use 8-bit T1 fonts
\usepackage{hyperref}       % hyperlinks
\usepackage{url}            % simple URL typesetting
\usepackage{booktabs}       % professional-quality tables
\usepackage{amsfonts}       % blackboard math symbols
\usepackage{nicefrac}       % compact symbols for 1/2, etc.
\usepackage{microtype}      % microtypography
\usepackage{xcolor}         % colors
\usepackage{graphicx}
\usepackage{amsmath}
\usepackage{multirow}
\usepackage{subcaption}
\usepackage{refcount}

\newcommand{\rev}[1]{#1}

\title{MODULE: A Mutual-Promoting Deep Unfolding Framework Towards Degradation-Robust Multi-modal Image Fusion}

\author{%
  Yunfei Deng\(^{1}\), \,
  Han Xu\(^{1}\)\thanks{Corresponding author.},  \,
  Jiayi Ma\(^{2}\), \,
  Guangcan Liu\(^{1}\) \\
  \(^{1}\)School of Automation, Southeast University, Nanjing, China \\
  \(^{2}\)School of Robotics, Wuhan University, Wuhan, China \\
  \texttt{xu\_han@seu.edu.cn, \{yunfeideng455, jyma2010\}@gmail.com,} \\
  \texttt{guangcanliu@seu.edu.cn}
}

\begin{document}

\maketitle

\begin{abstract}
Multi-modal image fusion is crucial for comprehensive scene representation, yet real-world degradations compromise its efficacy, necessitating degradation-robust fusion paradigms. Existing degradation-robust multi-modal image fusion methods are often hindered by cascaded sub-optimal solution or black-box opacity. To address these challenges, we propose a theory-inspired mutual-promoting deep unfolding framework. It reformulates degradation-robust image fusion as a joint optimization problem, utilizing the degradation model and modality generation mechanism to explicitly model both single-task priors and cross-task dependencies. By decomposing the complex multi-task optimization problem into task-related iterative subproblems, the framework establishes a bidirectional reciprocal information flow between restoration and fusion. This process is further unfolded into a multi-stage deep neural network, where each component explicitly corresponds to a specific mathematical operation. In our framework, the fused cross-modal prior actively regularizes the ill-posed restoration, while the purified modality features continuously refine the fusion output. It ensures a transparent architecture that combines the merits of both model-based and data-driven methodologies. Extensive experiments demonstrate that our paradigm achieves state-of-the-art performance while providing superior interpretability. The code is publicly
available at \url{https://github.com/Meskend/MODULE}.
\end{abstract}

\section{Introduction}

Due to complex imaging environments, single-modal images often fail to deliver a comprehensive representation of scenes. While visible images offer rich textures, they are compromised in low-light or occluded conditions. Conversely, infrared images can highlight thermal targets but lack fine-grained details. To bridge this gap, multi-modal image fusion (MMIF) aims to integrate complementary information to synthesize a single comprehensive representation. It is widely applied in diverse domains, \textit{e.g.}, autonomous driving~\cite{drive}, intelligent security~\cite{security}, and military surveillance~\cite{military}.

In recent years, learning-based multi-modal image fusion (MMIF) methods have achieved remarkable advancements, which can be categorized into methods based on CNN~\cite{cnn1, cnn2}, GAN~\cite{gan1, gan2}, Transformer~\cite{trans1,trans2}, Mamba~\cite{mamba1,mamba2}, and Diffusion~\cite{diff1,diff2,diff3}. However, they are fundamentally tailored to fusion, inherently assuming that source images are perfectly clean, overlooking complex degradations (\textit{e.g.}, noise, blur, and low light) that frequently occur under challenging imaging conditions. Consequently, their performance deteriorates significantly when deployed in practical scenarios, underscoring the urgent need for degradation-robust multi-modal image fusion.

Fundamentally, degradation-robust image fusion is not a sequential combination of isolated tasks, as restoration and fusion share a profound intrinsic dependency. High-quality restoration provides reliable features for fusion, while the fused output offers cross-modal guidance regularizing ill-posed restoration. As a result, the clean scene representation $z$ and the final fused result $x_f$ should be simultaneously estimated. This inherent connection naturally abstracts the degradation-robust fusion task into a joint optimization problem:
\begin{equation}
\min_{z, x_f} \mathcal{L}(z, x_f) = \mathcal{L}_{res}(z) + \mathcal{L}_{fus}(x_f) + \mathcal{L}_{couple}(z, x_f),
\end{equation}
where $\mathcal{L}_{res}(z)$ and $\mathcal{L}_{fus}(x_f)$ govern the single-task priors, and $\mathcal{L}_{couple}(z,x_f)$ explicitly encodes this intrinsic task dependency as a cross-task prior.

To tackle this challenging degradation-robust image fusion task, recent pioneering works have emerged, which can be broadly categorized into two main paradigms: cascaded methods and end-to-end black-box methods. However, when evaluated against the aforementioned theoretical yardstick, both paradigms exhibit inherent limitations when encoding task priors into their architecture design.

\begin{figure}[!t]
     \centering
     \includegraphics[width=\linewidth]{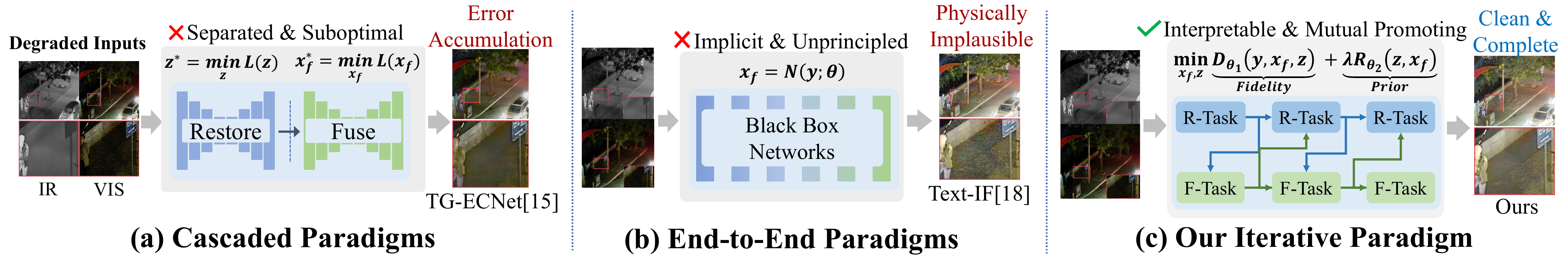}\\
     \vspace{-0.1cm}
     \caption{Comparison of existing degradation-robust image fusion paradigms. (a) Cascade paradigms suffer from biased task prior, causing error accumulation. (b) End-to-end paradigms abort the task-specific prior, leading to physically implausible results. (c) Our iterative paradigm is guided by task prior in optimization objectives, resulting in clean and comprehensive results. }
     \label{fig:first}
\vspace{-0.3cm}
\end{figure}

Cascaded methods, whether concatenating independent algorithms or decoupling processes sequentially within a single network~\cite{drmf, omnifuse, tgecnet, urfusion}, inherently adopt a biased prior encoding strategy, as Fig.~\ref{fig:first} (a) shows. These paradigms can be interpreted as solving the joint objective through a one-step sequential decomposition. By simply minimizing the single-task restoration term $\mathcal{L}_{res}(z)$ and completely ignoring the coupling term $\mathcal{L}_{couple}(z, x_f)$ at the initial stage, their final solution $(z^*, x_f^*)$ frequently leaves a non-zero gradient for the coupling term (\textit{i.e.}, $\nabla_{z}\mathcal{L}_{couple}(z^*, x_f^*) \neq 0$). This implies that cascaded strategies violate the mathematical optimality condition, trapping the solution in a suboptimal state. This mathematical suboptimality translates directly into a biased structural prior within the network architecture. By hard-coding a unidirectional information flow, the architecture forces the model to learn a biased mapping that prohibits exploring better solutions in the joint space. Consequently, this causes irreversible error accumulation: any degradation artifacts or over-smoothed details generated initially cannot be rectified in the subsequent fusion stage.

End-to-end black-box methods~\cite{Text-IF, deno} completely abandon the explicit modeling of the joint constraints and coupling rules formulated above, as Fig.~\ref{fig:first} (b) shows.  These task-prior-agnostic architectures favor the heuristic stacking of generic operators such as convolutions or Transformers, instantiating neither a specific degradation model nor a dedicated coupling mechanism. With task dependencies left to be implicitly inferred purely from training data, the coupled processes degenerate into an uninformed implicit mapping, forcing the model to blindly search an overwhelmingly vast hypothesis space. Ultimately, this lack of architectural regularization not only renders inference uninterpretable, but also provokes mutual interference between degradation removal and feature integration, frequently yielding physically implausible results that violate the intrinsic task dependency.

These limitations of both paradigms point to a common imperative: the network architecture should not be a heuristic assembly of generic operators, nor should it instantiate a biased decomposition of the joint task. Instead, it should faithfully reflect the task-related prior encoded in the optimization objective itself. A principled way to achieve this is to derive the architecture directly from an iterative algorithm that solves the joint objective. This approach, known as deep unfolding, provides an architectural scaffold that is both interpretable and task-prior-guided, guiding the restoration and fusion processes to evolve jointly toward a mutually consistent solution.
% To overcome the limitations of cascaded and black-box paradigms, it is imperative to establish a \textbf{bidirectional} and \textbf{principled} interaction between restoration and fusion to break single-task bottlenecks. By utilizing fused information as cross-modal priors, the ill-posed restoration process can be regularized to recover details that are unrecoverable in isolation. It establishes a synergistic feedback loop, mutually reinforcing two tasks. An explicit architectural prior is equally critical. Unlike black-box mappings, constructing the architecture directly upon a task-specific prior establishes a structural consistency among the fused image, restored images, and raw inputs. It encourages the output to be physically plausible rather than merely a perceptual approximation.

Guided by these motivations, we propose MODULE, a Mutual-prOmoting Deep UnfoLding framEwork that synergistically integrates restoration and fusion, as Fig.~\ref{fig:first} (c) shows. First, we reformulate the restoration and fusion processes into a single joint optimization problem. This mathematical formulation reveals the intrinsic interdependence between two tasks, laying the foundation for algorithmic transparency. Second, the complex multi-task optimization is decomposed into a series of iterative, single-task subproblems, where the solution of each subproblem is guided by others. The iterative nature establishes a reciprocal information flow, steering the system toward a mutually consistent and physically plausible solution. Finally, we unfold the optimization process into a multi-stage deep network architecture. Constructed directly upon an optimization-based task-specific prior, MODULE not only inherits the interpretability of model-based methods but also leverages the representational capacity of networks for superior performance. Our contributions are as follows:

1. We reformulate degradation-robust image fusion as a joint optimization problem from a generative perspective. By explicitly modeling the dependency  between restoration and fusion, our method effectively overcomes the limitations of cascaded suboptimal solutions and black-box opacity.

2. We decompose the complex multi-task optimization problem into a series of iterative, single-task subproblems, where the optimization of each is dynamically guided by the others. This iterative nature facilitates a bidirectional information flow, fostering mutual promotion that drives the system toward a mutually consistent and physically plausible solution.

3. We unfold the iterative optimization process into a multi-stage deep neural network. Each unfolded stage explicitly corresponds to a specific physical operation. It provides a transparent and interpretable architecture, seamlessly combining the advantages of model-based and data-driven approaches.

\section{Related Work}
% \textbf{Degradation-Robust Image Fusion.} Achieving robust multi-modal image fusion against complex degradations remains a primary challenge. One mainstream paradigm is to integrate restoration and fusion into a cascaded framework. For example, URFusion~\cite{urfusion} extracts degradation-invariant features for content fusion and leverages high-quality priors for appearance enhancement of the fusion result. DRMF~\cite{drmf} utilizes twin diffusion models for explicit restoration before fusing intermediate results during sampling. TG-ECNet~\cite{tgecnet} employs a mixture-of-experts architecture for both degradation removal and fusion. The other utilizes black-box networks to implicitly learn robust mappings through massive data. Text-IF~\cite{Text-IF} trains an end-to-end network to generate high-quality fused images and integrates language guidance to handle multiple degradations. Deno-IF~\cite{deno} utilizes low-rank decomposition to separate clean components from noise as training guidance and trains a joint denoising and fusion network. Overall, while effective, these methods often fail to bridge the gap between modular task separation and end-to-end interpretability.
\textbf{Degradation-Robust Image Fusion.} Achieving robust fusion against complex degradations remains challenging. Existing methods generally follow two paradigms. The cascaded paradigm integrates restoration and fusion; for example, URFusion~\cite{urfusion} relies on high-quality priors to enhance degradation-invariant features, while DRMF~\cite{drmf} utilizes twin diffusion models for explicit pre-restoration. Conversely, the end-to-end paradigm implicitly learns robust mappings. Within this category, Text-IF~\cite{Text-IF} adopts language guidance to handle diverse degradations, and Deno-IF~\cite{deno} separates clean components via low-rank decomposition. Yet, these methods fail to capture the inherent coupling between restoration and fusion, thereby preventing the two tasks from mutually regularizing and enhancing each other.

% \textbf{Deep Unfolding Networks.} Deep unfolding networks (DUNs) bridge the gap between model-based optimization and deep learning by unfolding iterative algorithms into layer-wise architectures. These frameworks have been extensively explored across a wide range of image reconstruction and restoration tasks. For example, in the field of compressive sensing, ISTA-Net~\cite{ista} translates shrinkage-thresholding iterations into a sparse nonlinear network. UFC-Net~\cite{ufc} refines inter-stage feature extraction via fixed-point optimization to enhance reconstruction performance. For image deblurring, DGUNet~\cite{dgu} integrates a flexible gradient descent module to handle dynamic degradations. To achieve multi-task restoration, VLU-Net~\cite{vlu} incorporates language prompts into the unfolding framework to guide recovery process. Despite these advancements, existing DUNs remain primarily limited to processing single-modal images. Consequently, there is a need for further research to advance DUNs in integrating multi-modal images and tackling multiple tasks concurrently.
\textbf{Deep Unfolding Networks.} Deep unfolding networks (DUNs) bridge model-based optimization and deep learning via layer-wise iterative architectures. They are extensively explored in image restoration. ISTA-Net~\cite{ista} unfolds shrinkage-thresholding iterations for compressive sensing. DGUNet~\cite{dgu} integrates gradient descent modules for image deblurring. VLU-Net~\cite{vlu} uses language prompts for multi-task recovery. IterRF~\cite{iterrf} unfolds a unified registration–fusion optimization into alternating subproblems, enabling iterative bidirectional refinement of both tasks. Despite these advances, existing DUNs mainly focus on single-modal restoration or registration–fusion coordination, without explicitly addressing the coupled restoration and fusion of multiple modalities.

\section{Method}
\subsection{Problem Formulation}
In practical deployments, multi-modal sensors are susceptible to environmental variations, hardware imperfections, and operational inaccuracies. With $i,v$ representing infrared and visible modalities, we can formulate the degraded source images based on the physical imaging process as:
\begin{equation}
y_m=D_mz_m+n_m, \ m\in{\{i,v\}},\label{eq_observation}
\end{equation}
where $y_m$ is the observed degraded source image. $z_m$ is the corresponding high-quality image. $D_m$ is the complex degradation operator (\textit{e.g.}, blurring, low light). $n_m$ represents noise.

\begin{figure*}[!t]
     \centering
     \includegraphics[width=\linewidth]{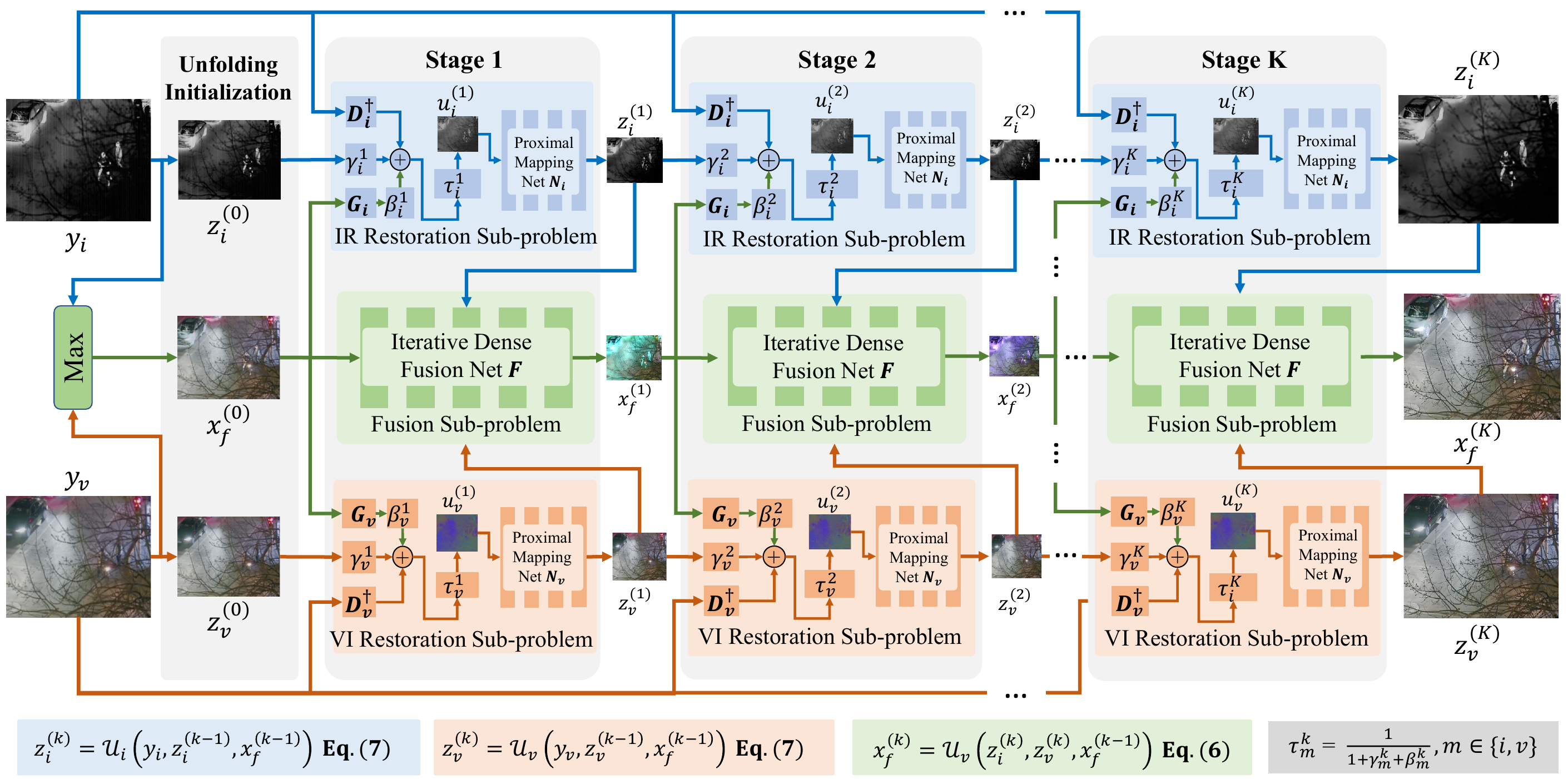}\\
     \vspace{-0.05in}
     \caption{Overall network architecture of the proposed MODULE.}
     \label{fig:method}
\vspace{-0.4cm}
\end{figure*}

Eq.~\eqref{eq_observation} is a highly ill-posed inverse problem. Restoring $z_m$ solely from the degraded observation $y_m$ is inherently limited, hampering restoration performance. Fortunately, source images in the image fusion task can provide complementary information about the same scene. The auxiliary modality offers a cross-modal prior to regularize the solution space, facilitating better restoration. However, challenges persist due to the significant discrepancy between different modalities. The irrelevant information in the auxiliary modality will interfere with the learning of correct mapping.

Thus, the core imperative is to maximize the guidance of cross-modal priors while suppressing cross-modal interference. It aligns with the goal of image fusion, \textit{i.e.}, bridging modality gaps and aggregating informative features into a fused image $x_f$. The fusion process distills a purified, cross-modal prior, making $x_f$ a comprehensive representation. Under this framework, two modalities are naturally modeled as distinct spectral projections of a unified prior. Thus, the high-quality source images can be generated by projecting the prior onto respective domains. From a generative perspective, the relationship is mathematically defined as:
\begin{equation}
z_m=G_mx_f, \ m\in{\{i,v\}},\label{eq_generation}
\end{equation}
where $G_m$ \rev{denotes a learnable modality-specific projection mapping, which projects} the fused image back into source domains.

When linking the fusion prior to restored images, it raises a question: how to get $x_f$ at first? Ideally, a high-quality $x_f$ should be derived from $z_m$ while generating $z_m$ conversely relies on $x_f$ and $y_m$. It reveals a circular dependency: estimating $x_f$ requires $z_m$, and estimating $z_m$ requires $x_f$. The interdependence cannot be resolved by sequentially performing fusion followed by restoration or vice versa.
To this end, we unify fusion and restoration into a single, jointly optimized framework. It allows the two processes to mutually inform and constrain each other.

Mathematically, this problem can be formulated as a Maximum a Posteriori (MAP) estimation problem, \textit{i.e.}, maximizing the posterior distribution $p\left(x_f, z_i, z_v\middle| y_i,y_v\right)$. Under the common assumption, $n_m$ can be reformulated as the quadratic data fidelity term through negative log-likelihood of MAP estimation. We transform the hard constraint of Eq.~\eqref{eq_generation} into an optimizable soft penalty via the quadratic penalty method. Thus, $x_f,z_m$ can be obtained by minimizing the following objective:
\begin{equation}
\begin{aligned}
\min_{x_f, z_i, z_v}  \sum\nolimits_{m\in\{i,v\}}\frac{1}{2} \| y_m - D_mz_m \|_F^2 + \lambda_m R_m(z_m)+ \frac{\beta_m}{2} \| z_m - G_mx_f \|_F^2 + \lambda_f R_f(x_f),
\end{aligned}\label{eq_forward_objective}
\end{equation}
where $\| y_m - D_mz_m \|_F^2$ and $\| z_m - G_mx_f \|_F^2$ are fidelity terms. $R_m(z_m)$ and $R_f(x_f)$ are regularization terms. $R_m$ imposes prior constraints on the high-quality source images. $R_f$ constrains the fused image. $\lambda_m$, $\beta_m$, and $\lambda_f$ control the trade-off.
This formulation narrows the solution space to a physically valid manifold, ensuring that recovered details are structurally consistent with the scene.

\rev{The original objective in Eq.~\eqref{eq_forward_objective} measures data fidelity in the observation space, where $D_m$ may be severely ill-conditioned or non-invertible: the data term is nearly flat along the weakly observed directions and leaves the null-space components unconstrained, so the estimate there is driven by noise and by the regularizer alone. We therefore introduce the Moore--Penrose pseudo-inverse $D_m^\dagger$, which is well defined for any bounded linear $D_m$ even when its analytical form is unavailable, and measure the residual in the restoration space instead:}
\begin{equation}
\begin{aligned}
\min_{x_f, z_i, z_v}  \sum\nolimits_{m \in \{i,v\}} \frac{1}{2} \| D_m^{\dagger} y_m - z_m \|_F^2 + \lambda_m R_m(z_m) + \frac{\beta_m}{2} \| z_m - G_m x_f \|_F^2 + \lambda_f R_f(x_f).
\end{aligned}
\label{surrogate}
\end{equation}
\rev{As analyzed in Appendix~\ref{app:theory}, this reformulation yields an isotropic data term and explicitly penalizes the null-space component of $z_m$: among all $z_m$ that explain the observation equally well, it selects the minimum-norm one and leaves the remaining null-space content to the prior. The ill-conditioned valley of the original data term thus becomes a well-conditioned one, and noise-driven variations in the unobservable directions are suppressed. It is a deliberate change of the objective rather than an equivalent rewriting of it.}

\rev{In practice, $D_m$ is unknown or lacks an explicit form; and even when available, a fixed linear inverse amplifies noise and cannot fill the unobservable directions. We therefore employ a prompt-driven neural network to learn a pseudo-inverse-inspired mapping that draws on image priors and cross-modal context to supply plausible structure where the observation carries none. Since it takes over the role of $D_m^\dagger y_m$ in Eq.~\eqref{surrogate}, we retain this notation below without assuming any exact pseudo-inverse properties. Unlike a standalone restoration network, its output is consumed by the fusion branch and optimized under the joint objective, so it preserves what fusion needs rather than merely minimizing a restoration loss. We unfold the joint optimization into alternating subproblems and implement them as a deep network (Fig.~\ref{fig:method}).}

\subsection{Iterative Subproblem Decomposition}
To solve this optimization problem, we use half-quadratic splitting (HQS) to decompose the complex multivariate ($x_f,z_i,z_v$) optimization problem into a series of iterative optimization sub-problems. The fusion sub-problem reconstructs the fused image \(x_f^{(k)}\) from the current refined modal-specific images and previous fusion estimate, creating restoration-to-fusion promotion. With $k$ denoting k-th optimization step, $x_f$-related (fusion) sub-problem can be formulated as:
\begin{equation}
x_f^{(k)} = \arg\min_{x_f} \frac{1}{2} \sum\nolimits_{m \in \{i,v\}} \beta_{m} \| z_m^{(k)} - G_m x_f \|_F^2 + \lambda_f R_f(x_f).
\end{equation}
The restoration sub-problem estimates \(z_m\) from both observations and previously iterated fused result. $z_m$-related (restoration) sub-problem can be formulated as:
\begin{equation}
z_m^{(k)} = \arg\min_{z_m}  \frac{1}{2} \| D_m^{\dagger} y_m - z_m \|_F^2 + \lambda_m R_m(z_m)  + \frac{\beta_{m}}{2} \| z_m - G_m x_f^{(k-1)} \|_F^2,  m \in \{i, v\}.
\end{equation}
To resolve \(z_m\)-related sub-problem, we perform HQS to decouple the fidelity from the implicit image prior. By introducing auxiliary variable \(u\) to approximate \(z\), \(u\)-subproblem is constructed as:
\begin{equation}
u_m^{(k)}=\arg \min_{u_m} \frac{1}{2} \|D_m^{\dagger} y_m-u_m\|_F^2+\frac{\beta_{m}}{2}\|u_m-G_m x_f^{(k-1)} \|_F^2\\
+ \frac{\gamma_{m}^{(k)}}{2} \| u_m - z_m^{(k-1)}\|_F^2,
\end{equation}
where $\gamma_m^{(k)}$ is a penalty parameter. \(z\)-subproblem is constructed as:
\begin{equation}
z_m^{(k)} = \arg\min_{z_m}  \lambda_m R_m(z_m) + \frac{\gamma_{m}^{(k)}}{2} \| z_m - u_m^{(k)} \|_F^2.
\end{equation}
$u_m$ enforces physical fidelity. $z_m$ yields prior-enhanced reconstruction.

\subsection{Optimization-Inspired Architecture Realization}
To enable data-driven learning while maintaining interpretability, we instantiate each sub-problem as a dedicated network module.
Mathematically, the \(u\)-subproblem admits a closed-form solution as:
\begin{equation}
u_m^{(k)} = \frac{D_m^{\dagger} y_m + \beta_m G_m x_f^{(k-1)} + \gamma_m^{(k)} z_m^{(k-1)}}{1 + \beta_m + \gamma_m^{(k)}}.
\end{equation}
It acts as a fusion hub. Restoration is guided by fused \(G_m x_f^{(k-1)}\) for fusion-to-restoration promotion.

\(z\)-subproblem is equivalent to a proximal mapping characterized by prior \(R_m\). As a high-dimensional prior is difficult to model, we instantiate it with a network \(\mathcal{N}_m\) to learn the manifold of high-quality images:
\begin{equation}
z_m^{(k)} = \mathcal{N}_m(u_m^{(k)}; \theta_k) \approx \text{prox}_{\frac{\lambda_m}{\gamma_m^{(k)}}}(u_m^{(k)}).
\label{eq:proximal_mapping}
\end{equation}
The fusion sub-problem aims to reconstruct the comprehensive scene \(x_f\) from the refined modality-specific images \(z_i^{(k)},z_v^{(k)}\). Instead of splitting \(x_f\), we employ a fusion network \(\mathcal{F}\) to learn high-dimensional feature aggregation directly. It avoids hand-crafted fusion rules and enables the model to distill ``restoration-critical'' features from \(z_m^{(k)}\), realizing restoration-to-fusion promotion as:
\begin{equation}
x_f^{(k)} = \mathcal{F}(z_i^{(k)}, z_v^{(k)}, x_f^{(k-1)}).
\end{equation}
\textbf{Initialization.} We initialize $z_m^{(0)} = y_m$. For the fused image, we adopt a pixel-wise maximum selection strategy, \textit{i.e.}, $x_f^{(0)} = \max(y_i, y_v)$ to preserve salient thermal targets and structural textures across modalities, establishing a robust starting point for fast convergence during unfolding iterations.

\textbf{Network Instantiation.} We further instantiate the mathematical operators with specialized deep neural networks. The architecture is categorized into Operator Simulation Networks (instantiating $D_m^\dagger$ and $G_m$) and Solver Simulation Networks (instantiating $\mathcal{N}_m$ and $\mathcal{F}$). These networks are organized into $K$ stages, sharing both the same architecture and learnable parameters.

\textit{Operator Simulation Networks.}
%These networks parameterize the physical degradation $D_m^\dagger$ and domain mapping operators $G_m$ to ensure physical consistency. First, inspired by PromptIR~\cite{promptir},
$D_m^{\dagger}$ is realized by a prompt-driven U-Net constructed by NAFBlocks~\cite{chen2022simple}. Each block comprises a Prompt Generation Module (PGM) and a Prompt Interaction Module (PIM). PGM generates degradation-aware prompts from input features $F$. First, to capture global degradation context, we derive a content-adaptive weight vector $w$ with global average pooling and gating mechanism as $w = \sigma(\text{Conv}_{gate}(\text{GAP}(F)))$, where $\sigma$ is Sigmoid. Then, $w$ modulates a learnable prompt bank $\mathbf{P}_{base}$, followed by a convolutional projection to obtain the specific prompt $P = \text{Conv}_{up} \left( w \odot \mathbf{P}_{base} \right)$. Then, PIM facilitates the interaction between features and prompts. By concatenating $F$ and $P$, a NAFBlock refines features as $\hat{F} = \text{Conv}_{out} ( \text{NAFBlock} ([F, P]))$.
$G_m$ is instantiated as a network of stacked Restormer~\cite{restormer} blocks. With multi-head self-attention mechanism, $G_m$ captures long-range dependencies, projecting the latent fused representation $x_f$ back into modal-specific feature spaces while preserving spatial consistency. \rev{It should be emphasized that $G_m$ is not required to reconstruct the full modality image; its role is to provide a modality-specific projection that transfers cross-modal structural priors to the restoration branch, and the hard relation $z_m=G_m(x_f)$ is relaxed into the soft consistency constraint $L_p$.}

\textit{Solver Simulation Networks.} Restoration solver $\mathcal{N}_m$, corresponding to $z$-subproblem, is constructed as a residual network embedded with squeeze-and-excitation blocks~\cite{se}. Through channel-wise attention, $\mathcal{N}_m$ dynamically recalibrates feature responses, emphasizing thermal targets and texture details.
The fusion solver $\mathcal{F}$, corresponding to $x_f$-subproblem, adopts an asymmetric encoder-decoder. The encoder employs dense connections to maximize feature reuse. %Let $f_0 = [z_i^{(k)}, z_v^{(k)}]$ be the concatenation of the refined single-modal images.
%\begin{equation}
%f_l = H_l([f_0, f_1, \dots, f_{l-1}]).
%\end{equation}
The decoder utilizes plain convolutions. For restoration-to-fusion promotion, we use a residual interaction from previous stage to learn incremental refinement of scene representation, ensuring stability and high-frequency preservation. $\beta_m,\gamma_m$ are stage-wise learnable to adaptively balance data fidelity and prior regularization.

\subsection{Loss Function}
The overall loss function consists of two components: i) fusion loss $\mathcal{L}_{fuse}$ to regulate the fused image $x_f$, and ii) consistency loss $\mathcal{L}_{cons}$ to constrain the intermediate variables in iterative unfolding process, finally formulated as $\mathcal{L}_{total} = \mathcal{L}_{fuse} + \mathcal{L}_{cons}$.

\textbf{Fusion Loss}. To ensure that $x_f$ preserves salient targets, fine textures, and natural color, we define $\mathcal{L}_{fuse}$ as a combination of intensity, gradient, and color constraints as:
\begin{equation}
\mathcal{L}_{fuse} = \alpha_1 \mathcal{L}_{int} + \alpha_2 \mathcal{L}_{grad} + \alpha_3 \mathcal{L}_{color},
\end{equation}
where $\alpha_1,\alpha_2,\alpha_3$ are hyperparameters to control trade-off.

The intensity loss $\mathcal{L}_{int}$ measures the intensity consistency between $x_f$ and high-quality images $I_{v}, I_{i}$ as $\mathcal{L}_{int} = \frac{1}{HW} \Vert x_f - \max(I_{v}, I_{i}) \Vert_1$. $H,W$ are height and width. With the Sobel operator $\nabla(\cdot)$, the gradient loss $\mathcal{L}_{grad}$ constrains the alignment of edges as $\mathcal{L}_{grad} = \frac{1}{HW} \Vert \nabla x_f - \max(\nabla I_{v}, \nabla I_{i}) \Vert_1$. The color loss $\mathcal{L}_{color}$ constrains color consistency as $\mathcal{L}_{color} = \frac{1}{HW} \Vert \mathcal{F}_{CbCr}(x_f) - \mathcal{F}_{CbCr}(I_{vis}) \Vert_1$ where $\mathcal{F}_{\text{CbCr}}(\cdot)$ extracts the Cb and Cr chrominance channels.

\textbf{Consistency Loss.} This loss enforces the physical and mathematical alignment across the restoration and fusion tasks within all $K$ stages:
\vspace{-0.1cm}
\begin{equation}
\mathcal{L}_{cons} = \sum\nolimits_{k=1}^{K} \sum\nolimits_{m \in \{i,v\}}  \lambda_{z}^{(k)} \mathcal{L}_{z}^{(k)} + \lambda_{u}^{(k)} \mathcal{L}_{u}^{(k)} + \lambda_{p}^{(k)} \mathcal{L}_{p}^{(k)}.
\end{equation}
% \vspace{-0.1cm}
Restoration fidelity loss $\mathcal{L}_{z}^{(k)} = \|z_m^{(k)} - I_m\|_1$ supervises $z_m$ to high-quality domain. With hyperparameter $\eta$, restoration loss of final $z_m$ is $\mathcal{L}_{z}^{K} = \mathcal{L}_{z} + \eta(1-SSIM(z_m, I_m))$ to further preserve structure similarity in final iteration. Proximal alignment loss $\mathcal{L}_{u}^{(k)}=\|u_m^{(k)} - z_m^{(k)}\|_1$ ensures consistency between analytical solution $u_m$ (derived from physics-based sub-problem) and $z_m$, maintaining integrity of HQS decomposition. Projection consistency loss $\mathcal{L}_{p}^{(k)}=\|G_m(x_f^{(k)}) - z_m^{(k)}\|_1$ enforces structural alignment between projection of $x_f$ and restored image for fusion-restoration promotion.

\section{Experiments and Results}
\textbf{Datasets and Settings}. We perform experiments on LLVIP~\cite{llvip} and MSRS~\cite{cnn1} datasets. For each dataset, training and testing sets comprise 800 and 200 degraded–clean image pairs. Five types of degradations (infrared Gaussian noise, infrared stripe noise, visible Gaussian noise, visible low-light conditions, and visible defocus blur) are considered. \rev{Degradation parameters are randomly sampled per pair; details are in Appendix~\ref{app:degradation}.} %Each image pair is subjected to one of these five degradations.
Experiments are conducted on an NVIDIA A800 GPU with PyTorch. We use the Adam optimizer, setting the initial learning rate to 0.001 with linear decay. Images are randomly cropped into 64$\times$64. Epoch is 1500. Batch size is 24. $K$ is 3. $\alpha_1$, $\alpha_2$, $\alpha_3$ are 6, 8, 12. $\lambda_u$ and $\lambda_p$ are 1.5 and 2.0 and increase by 0.1 per iteration. $\lambda_z$ is 1 for the first two iterations and 4 for final iteration. $\eta$ is 1. Hyperparameters are empirically determined.

\begin{figure*}[!t]
     \centering
        \includegraphics[width=\linewidth]{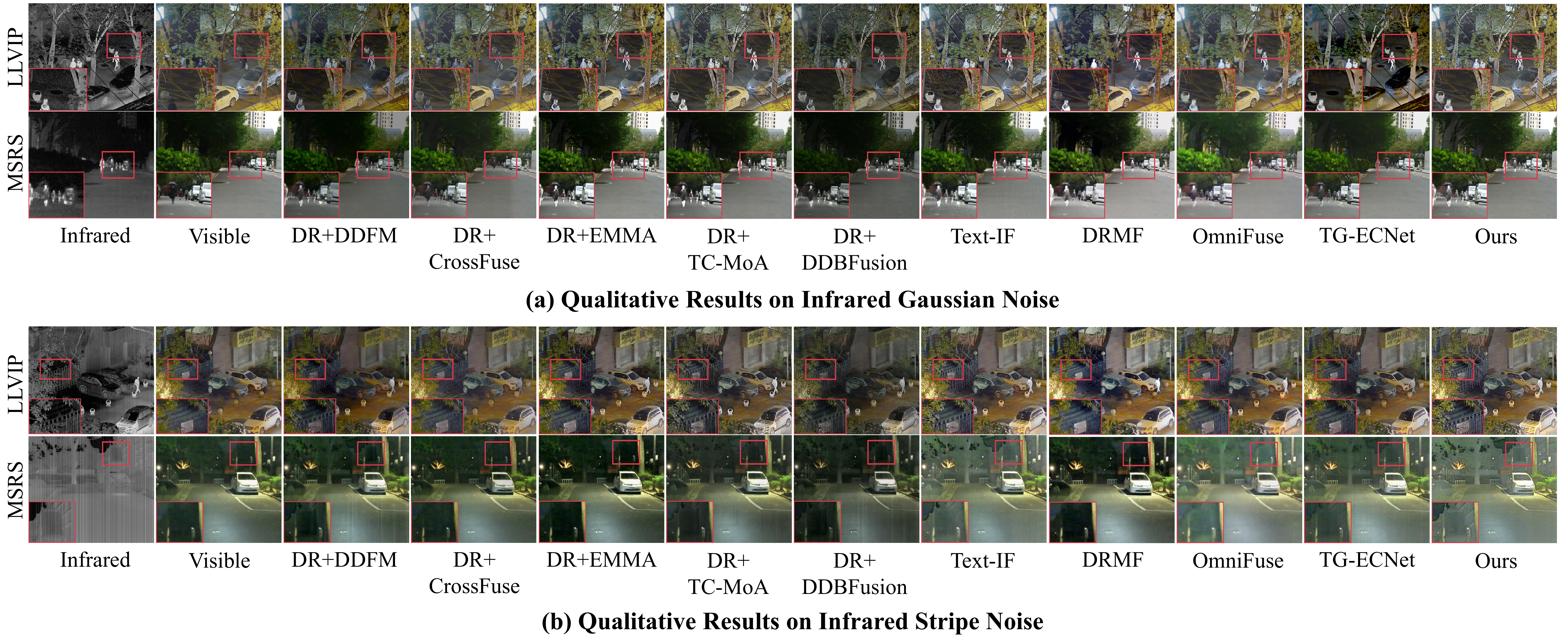}\\
        \vspace{-0.05in}
     \caption{Qualitative results when the infrared image suffers from degradations. }
     \label{fig:ir}
\end{figure*}

\textbf{Multi-modal Image Fusion in Degraded Scenarios}. Competitors include DDFM~\cite{diff1}, CrossFuse~\cite{crossfuse}, EMMA~\cite{emma}, TC-MoA~\cite{tcmoa}, DDBFusion~\cite{DDBFusion}, Text-IF~\cite{Text-IF}, DRMF~\cite{drmf}, OmniFuse~\cite{omnifuse}, and TG-ECNet~\cite{tgecnet}. For typical fusion methods, degraded infrared images are preprocessed by degradation-robust diffusion model in DRMF. Degraded visible images are preprocessed by an all-in-one restoration method VLU-Net~\cite{vlu}. \rev{Note that Text-IF, DRMF, OmniFuse and TG-ECNet remove degradations and fuse end-to-end.}

\begin{figure*}[!t]
     \centering
        \includegraphics[width=\linewidth]{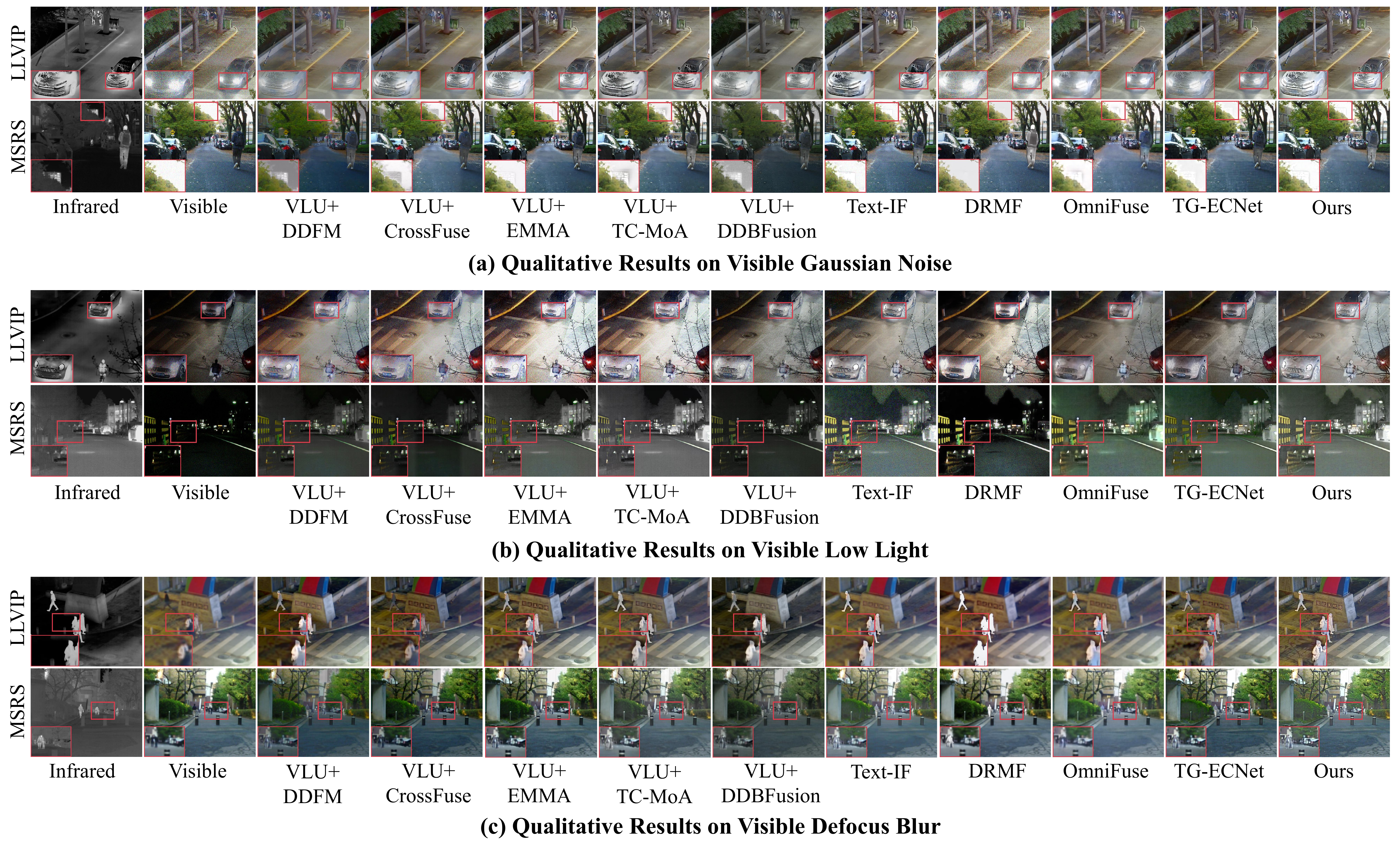}\\
        \vspace{-0.05in}
     \caption{Qualitative results when the visible image suffers from degradations. }
     \label{fig:vi}
\end{figure*}

\textbf{\textit{i) Qualitative Comparison}.} Results when the infrared or visible image suffers from degradations are shown in Figs.~\ref{fig:ir}-\ref{fig:vi}, respectively. Our framework demonstrates integrated advantages in restoration and fusion via its coupled modeling and bidirectional guidance. For restoration, the fused image serves as the cross-modal prior that removes degradations, and the physically plausible modeling keeps all variables reciprocally regularized; in contrast, competing methods leave noticeable artifacts or residuals owing to their cascaded or black-box nature. For fusion, the high-fidelity restored results supply complementary information from both modalities, and the generative projection constraint of the unfolding network preserves salient thermal targets and fine textures.

\textbf{\textit{ii) Quantitative Comparison.}} Evaluation metrics include SCD~\cite{scd}, VIF~\cite{VIF}, $Q_{abf}$~\cite{Qabf} and SSIM~\cite{SSIM}. Preprocessing methods are referred to as Restoration Methods (RM) for simplicity. In Tab.~\ref{tab:quantitative_horizontal}, existing methods are constrained by cascaded pipelines or black-box designs and fail to disentangle valid scenes. In contrast, our method \rev{achieves best or second-best results on most metrics} across datasets and degradations. Per-degradation results are given in Appendix~\ref{app:quantitative}.

\begin{table*}[!t]
  \centering
  \caption{Quantitative results on degraded infrared and visible image fusion ({\color{red}best}, {\color{blue}second-best}).}
  \label{tab:quantitative_horizontal}
  \vspace{-0.05in}
  \resizebox{\linewidth}{!}{
  \renewcommand\arraystretch{1.1}
    \begin{tabular}{l@{\,~}|c@{\,~~~}c@{\,~~~}c@{\,~~~}c@{\,}|c@{\,~~~}c@{\,~~~}c@{\,~~~}c@{\,}|
    c@{\,~~~}c@{\,~~~}c@{\,~~~}c@{\,}|c@{\,~~~}c@{\,~~~}c@{\,~~~}c@{\,}}
      \toprule
      \textbf{Datasets} & \multicolumn{8}{c|}{\textbf{LLVIP Dataset}} & \multicolumn{8}{c}{\textbf{MSRS Dataset}} \\
      \cmidrule(lr){0-9} \cmidrule(lr){10-17}
      \textbf{Degradations} & \multicolumn{4}{c|}{\textit{Infrared Degradations}} & \multicolumn{4}{c|}{\textit{Visible Degradations}} & \multicolumn{4}{c|}{\textit{Infrared Degradations}} & \multicolumn{4}{c}{\textit{Visible Degradations}} \\
      \midrule
      Methods & SCD$\uparrow$ & VIF$\uparrow$ & $Q^{abf}$$\uparrow$ & SSIM$\uparrow$ & SCD$\uparrow$ & VIF$\uparrow$ & $Q^{abf}$$\uparrow$ & SSIM$\uparrow$ & SCD$\uparrow$ & VIF$\uparrow$ & $Q^{abf}$$\uparrow$ & SSIM$\uparrow$ & SCD$\uparrow$ & VIF$\uparrow$ & $Q^{abf}$$\uparrow$ & SSIM$\uparrow$ \\
      \midrule

      % 第一行：RM+DDFM
      RM+DDFM & 1.293 & 0.400 & 0.301 & 0.650 & \color{red}1.575 & 0.445 & 0.338 & \color{blue}0.683 & 1.282 & 0.480 & 0.289 & 0.527 & \color{red}1.354 & 0.654 & 0.300 & 0.667 \\

      % 第二行：RM+CrossFuse
      RM+CrossFuse & 0.754 & 0.355 & 0.307 & 0.651 & 0.831 & 0.420 & 0.286 & 0.530 & 0.963 & 0.487 & 0.353 & 0.669 & 0.959 & 0.518 & 0.276 & 0.528 \\

      % 第三行：RM+TC-MoA
      RM+TC-MoA & 1.181 & 0.414 & 0.374 & 0.699 & 1.287 & 0.478 & 0.399 & 0.663 & 1.293 & \color{blue}0.542 & \color{blue}0.374 & \color{blue}0.683 & 1.257 & \color{blue}0.765 & \color{blue}0.392 & \color{blue}0.760 \\

      % 第四行：RM+EMMA
      RM+EMMA & 1.283 & 0.361 & 0.333 & 0.663 & 1.374 & 0.447 & 0.330 & 0.602 & 1.353 & 0.516 & 0.362 & 0.662 & 1.315 & 0.613 & 0.337 & 0.628 \\

      % 第五行：RM+DDBFusion
      RM+DDBFusion & 1.255 & 0.359 & 0.246 & 0.607 & \color{blue}1.531 & 0.402 & 0.280 & 0.656 & 1.229 & 0.402 & 0.241 & 0.473 & 1.299 & 0.598 & 0.271 & 0.644 \\

      % 第六行：Text-IF
      Text-IF & \color{blue}1.322 & \color{red}0.464 & \color{blue}0.453 & \color{blue}0.739 & 1.266 & \color{blue}0.466 & \color{blue}0.422 & 0.648 & \color{blue}1.378 & 0.480 & 0.345 & 0.443 & 1.291 & 0.552 & 0.341 & 0.501 \\

      % 第七行：DRMF
      DRMF & 1.078 & 0.395 & 0.322 & 0.644 & 0.898 & 0.410 & 0.275 & 0.421 & 1.214 & 0.505 & 0.322 & 0.598 & 1.153 & 0.445 & 0.257 & 0.382 \\

      % 第八行：OmniFuse
      OmniFuse & 1.220 & 0.366 & 0.255 & 0.602 & 1.166 & 0.420 & 0.244 & 0.539 & 1.167 & 0.519 & 0.280 & 0.498 & 1.146 & 0.568 & 0.291 & 0.508 \\

      % 第九行：TG-ECNet
      TG-ECNet & 1.083 & 0.356 & 0.313 & 0.669 & 1.132 & 0.404 & 0.292 & 0.546 & 0.361 & 0.497 & 0.352 & 0.511 & 1.205 & 0.541 & 0.309 & 0.567 \\

      \midrule
      % 第十行：我们的方法 (Ours)
      \textbf{Ours} & \color{red}1.480 & \color{blue}0.455 & \color{red}0.462 & \color{red}0.773 & 1.448 & \color{red}0.575 & \color{red}0.498 & \color{red}0.831 & \color{red}1.389 & \color{red}0.621 & \color{red}0.395 & \color{red}0.788 & \color{blue}1.345 & \color{red}0.769 & \color{red}0.451 & \color{red}0.782 \\
      \bottomrule
    \end{tabular}
  }
\end{table*}

\begin{figure}[!t]
     \centering
        \includegraphics[width=\linewidth]{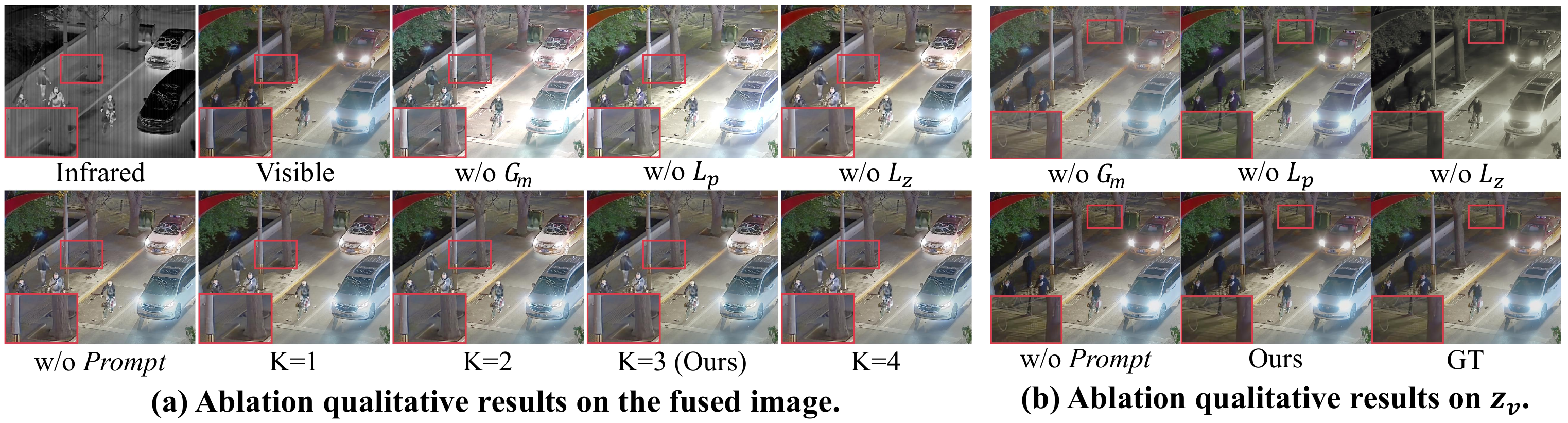}\\
        \vspace{-0.05in}
     \caption{Qualitative results of ablation study. }
     \label{fig:ablation}
\end{figure}

\textbf{Ablation Study}. \rev{We ablate the proposed components, the unfolding depth and the mutual promotion between the two branches. Removing $G_m$, $L_p$, $L_z$ or the prompt block in $D^\dagger$ one at a time causes overexposure, color shifts, texture distortion and slight texture loss, and degrades most metrics in Tab.~\ref{tab:ablation_by_degradation}; increasing the unfolding depth improves performance up to three stages and slightly degrades it at four while the training losses in Tab.~\ref{tab:kloss} (Appendix~\ref{app:ablations}) keep decreasing, so we adopt $K=3$. For restoration to fusion promotion, the fused output in Fig.~\ref{fig:ablation}(a) deteriorates whenever the restoration branch is ablated, and the corresponding drops in Tab.~\ref{tab:ablation_by_degradation} confirm that better restoration benefits fusion. For fusion to restoration promotion, disabling the feedback makes the intermediate visible restoration lose the color and structure of the low-light scene, whereas the full model recovers them from the intact infrared structure in Fig.~\ref{fig:ablation}(b); the convergence of the iterative variables in Tab.~\ref{tab:convergence} and the quantitative evidence in Appendix~\ref{app:ablations} show that fusion in turn guides restoration.}

\begin{figure}[!t]
    \centering
    \begin{minipage}[t]{0.58\linewidth}
        \centering
        \vspace{0pt}
        \captionof{table}{Ablation studies on the LLVIP dataset.}
        \vspace{-0.05in}
        \label{tab:ablation_by_degradation}
        \resizebox{1\linewidth}{!}{
        \renewcommand{\arraystretch}{1.25}
        \begin{tabular}{l@{\,~}|c@{\,~~}c@{\,~~}c@{\,~~}c@{\,~}|
        c@{\,~~}c@{\,~~}c@{\,~~}c@{\,~}}
    \toprule
    \multirow{2}{*}[-0.5ex]{\textbf{Variant}} & \multicolumn{4}{c|}{\textbf{Infrared}} & \multicolumn{4}{c}{\textbf{Visible}} \\
    \cmidrule(lr){2-5} \cmidrule(lr){6-9}
    & SCD $\uparrow$ & VIF $\uparrow$ & $Q^{abf} \uparrow$ & SSIM $\uparrow$ & SCD $\uparrow$ & VIF $\uparrow$ & $Q^{abf} \uparrow$ & SSIM $\uparrow$ \\
    \midrule
    w/o $G_m$             & 1.296 & 0.374 & 0.370 & 0.642 & 1.299 & 0.589 & 0.446 & 0.751 \\
    w/o $L_p$             & 1.284 & 0.384 & 0.368 & 0.631 & 1.319 & 0.596 & 0.440 & 0.706 \\
    w/o $L_z$             & 1.546 & 0.457 & 0.448 & 0.745 & 1.490 & 0.529 & 0.443 & 0.782 \\
    w/o Prompt            & 1.480 & 0.434 & 0.440 & 0.772 & 1.477 & 0.590 & 0.498 & {\color{red}0.874} \\
    \midrule
    $K=1$                 & 1.516 & 0.466 & 0.462 & 0.804 & 1.489 & 0.581 & 0.490 & 0.857 \\
    $K=2$                 & 1.518 & 0.467 & 0.460 & 0.801 & {\color{blue}1.503} & {\color{red}0.593} & {\color{red}0.499} & 0.872 \\
    \textbf{$K=3$ (Ours)} & {\color{red}1.545} & {\color{blue}0.471} & {\color{red}0.466} & {\color{red}0.815} & {\color{red}1.516} & {\color{blue}0.591} & {\color{blue}0.498} & {\color{blue}0.873} \\
    $K=4$                 & {\color{blue}1.539} & {\color{red}0.472} & {\color{blue}0.465} & {\color{blue}0.809} & 1.500 & 0.585 & 0.494 & 0.869 \\
    \bottomrule
    \end{tabular}}
    \end{minipage}
    \hfill
    \begin{minipage}[t]{0.41\linewidth}
        \centering
        \vspace{0pt}
        \includegraphics[width=\linewidth]{M3FD.pdf}\\
        \vspace{-0.1in}
        \captionof{figure}{Results on the unseen M3FD dataset under cross-modal degradation.}
        \label{fig:m3fd_qual}
    \end{minipage}

\end{figure}

\rev{\textbf{Generalization to Unseen and Compound Degradations.}} \label{sec:generalization} \rev{We test generalization under two conditions absent from training: an unseen dataset with both modalities degraded, and a compound degradation combining two corruptions of the visible image.} \rev{For the unseen setting, we evaluate on 100 M3FD pairs never involved in training and add Gaussian noise to both modalities, stripe noise to the infrared image, and low light with shot noise together with defocus blur to the visible image, with all strengths beyond the training ranges.} \rev{Fig.~\ref{fig:m3fd_qual} and Tab.~\ref{tab:gen}(a) show that most baselines either leave the noise and stripe patterns visible or suppress them at the price of washed-out thermal targets and lost visible textures, whereas MODULE keeps the thermal signatures sharp and the visible textures intact, ranking first on four of the five metrics and second on the remaining one; the only metric led by a baseline is PSNR, reflecting our emphasis on structural and perceptual fidelity over pixel-wise reconstruction under heavy simultaneous degradation.} \rev{Fig.~\ref{fig:compound_qual} in Appendix~\ref{app:compound} and Tab.~\ref{tab:gen}(b) show the same under compound degradation: MODULE preserves structural fidelity and gradient information even when luminance and sharpness cues are removed together.}

\begin{table*}[!t]
  \centering
  \caption{\rev{Evaluation on unseen and compound degradations} ({\color{red}best}, {\color{blue}second-best}).}
  \label{tab:gen}
  \vspace{-0.05in}
  \begin{subtable}[t]{0.49\linewidth}
    \centering
    \caption{\rev{Unseen M3FD, both modalities degraded beyond the training range.}}
    \label{tab:m3fd}
    \vspace{-0.05in}
    \resizebox{\linewidth}{!}{
    \renewcommand{\arraystretch}{1.1}
    \begin{tabular}{l@{\,~~}|c@{\,~~~}c@{\,~~~}c@{\,~~~}c@{\,~~~}c}
    \toprule
    Method & SCD $\uparrow$ & VIF $\uparrow$ & $Q^{abf} \uparrow$ & PSNR $\uparrow$ & SSIM $\uparrow$ \\
    \midrule
    DDFM      & {\color{blue}1.345} & 0.307 & 0.177 & 12.487 & 0.228 \\
    CrossFuse & 0.682 & 0.304 & 0.196 & 13.250 & 0.205 \\
    TC-MoA    & 1.034 & 0.328 & {\color{blue}0.222} & 13.686 & 0.236 \\
    EMMA      & 1.093 & 0.323 & 0.218 & 13.412 & 0.226 \\
    DDBFusion & 1.286 & 0.264 & 0.156 & 12.197 & 0.202 \\
    Text-IF   & 1.002 & 0.193 & 0.177 & 12.657 & 0.127 \\
    DRMF      & 0.812 & 0.263 & 0.188 & 12.905 & 0.174 \\
    OmniFuse  & 1.216 & {\color{blue}0.384} & 0.212 & 14.291 & {\color{blue}0.280} \\
    TG-ECNet  & 1.255 & 0.318 & 0.208 & {\color{red}14.968} & 0.235 \\
    \textbf{Ours} & {\color{red}1.512} & {\color{red}0.532} & {\color{red}0.348} & {\color{blue}14.345} & {\color{red}0.501} \\
    \bottomrule
    \end{tabular}}
  \end{subtable}
  \hfill
  \begin{subtable}[t]{0.49\linewidth}
    \centering
    \caption{\rev{Compound degradation on LLVIP (low light and defocus).}}
    \label{tab:compound}
    \vspace{-0.05in}
    \resizebox{\linewidth}{!}{
    \renewcommand{\arraystretch}{1.1}
    \begin{tabular}{l@{\,~~}|c@{\,~~~}c@{\,~~~}c@{\,~~~}c@{\,~~~}c}
    \toprule
    Method & SCD $\uparrow$ & VIF $\uparrow$ & $Q^{abf} \uparrow$ & PSNR $\uparrow$ & SSIM $\uparrow$ \\
    \midrule
    DDFM      & {\color{blue}1.254} & 0.376 & 0.168 & 14.013 & 0.372 \\
    CrossFuse & 0.845 & 0.291 & 0.083 & 14.199 & 0.284 \\
    TC-MoA    & 1.185 & {\color{red}0.484} & {\color{blue}0.195} & 14.013 & {\color{blue}0.422} \\
    EMMA      & 1.178 & 0.387 & 0.179 & 14.267 & 0.354 \\
    DDBFusion & 1.056 & 0.329 & 0.138 & 12.944 & 0.325 \\
    Text-IF   & 1.109 & 0.222 & 0.181 & 15.416 & 0.198 \\
    DRMF      & 0.838 & 0.229 & 0.114 & 13.974 & 0.189 \\
    OmniFuse  & 1.148 & 0.317 & 0.131 & {\color{red}16.364} & 0.359 \\
    TG-ECNet  & 1.187 & 0.367 & 0.175 & 14.322 & 0.347 \\
    \textbf{Ours} & {\color{red}1.333} & {\color{blue}0.398} & {\color{red}0.196} & {\color{blue}15.918} & {\color{red}0.488} \\
    \bottomrule
    \end{tabular}}
  \end{subtable}
\vspace{-0.3cm}
\end{table*}

\textbf{Extensive Experiments}. We conduct extensive experiments including real-world validation, interpretability and convergence analysis, high-level validation and complexity analysis.

\begin{figure}[!t]
     \centering
        \includegraphics[width=\linewidth]{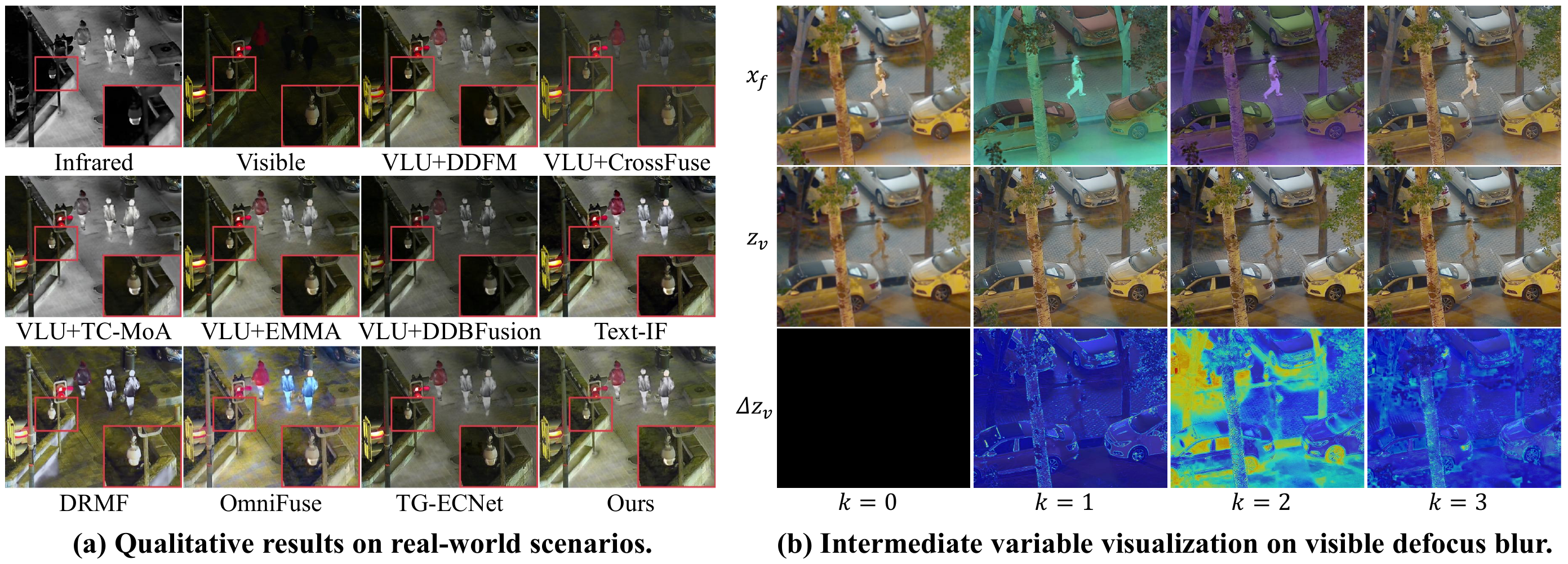}\\
        \vspace{-0.05in}
     \caption{Qualitative results of real-world validation and intermediate variable visualization.}
     \label{fig:visualization}
\end{figure}

\textbf{\textit{i) Real-world Validation}.} We select degraded images (scarce in datasets) from LLVIP and MSRS datasets, \rev{45 in total and none of which is involved in training}. Fig.~\ref{fig:visualization} (a) shows that our framework retains vital texture information while successfully enhancing the illumination of the visible image. Tab.~\ref{tab:real_convergence} demonstrates that our framework outperforms all competitors in non-reference metrics.

\textbf{\textit{ii) Interpretability and Convergence Analysis.}} Fig.~\ref{fig:visualization} (b) visualizes fused image $x_f$, restored visible image $z_v$ and iterative difference maps $\Delta z_v$ under visible defocus blur. The variables evolve in a coarse-to-fine manner, progressively eliminating ghosting and recovering details. It confirms our framework is transparent and task-guided. Tab.~\ref{tab:convergence} shows all intermediate variables converge to ideal targets during inference, indicating that our framework simulates the optimization algorithm.

\textbf{\textit{iii) High-level Validation and Complexity Analysis.}} We evaluate object detection performance in low-light scenarios on LLVIP with YOLOv11~\cite{yolo} as detector. Tab.~\ref{tab:combined_detection_complexity} reveals that MODULE achieves SOTA performance across all metrics as it can balance detail preservation and degradation suppression, achieving robustness in complex scenarios. \rev{In terms of computational cost, MODULE is lightweight in both parameters and FLOPs, so its multi stages do not sacrifice efficiency.}

\begin{table*}[!t]
  \centering
  \caption{Experiment on real-world images and convergence analysis of intermediate variables.}
  \label{tab:real_convergence}
  \vspace{-0.05in}
 \resizebox{1\linewidth}{!}{
 \renewcommand{\arraystretch}{1.1}
    \begin{tabular}{l@{\,~~~}|c@{\,~~~}c@{\,~~~}c@{\,~~~}c@{\,~~~}c@{\,~~~}c@{\,~~~}c@{\,~~~}c@{\,~~~}c@{\,~~~}c@{\,~~~}}
      \toprule
      Methods & DDFM & CrossFuse & TC-MoA & EMMA & DDBFusion & Text-IF & DRMF & OmniFuse & TG-ECNet & \textbf{Ours} \\
      \midrule
      NIQE $\downarrow$ & 4.360 & 4.253 & 4.200 & 4.668 & \textcolor{blue}{4.040} & 4.198 & 5.115 & 6.057 & 5.069 & \textcolor{red}{3.667} \\
      BRISQUE $\downarrow$ & 31.590 & 27.927 & 26.616 & 25.148 & \textcolor{blue}{23.288} & 28.663 & 41.255 & 47.784 & 24.457 & \textcolor{red}{22.356} \\
      MUSIQ $\uparrow$ & 43.046 & 44.482 & 45.368 & 44.385 & 43.332 & 46.212 & \textcolor{blue}{49.835} & 42.269 & 47.670 & \textcolor{red}{50.122} \\
      \bottomrule
    \end{tabular}
  }
\end{table*}

\begin{table*}[!t]
  \centering
  \caption{Convergence analysis under representative degradations ({\color{red}lowest}, {\color{blue}second-lowest}).}
  \label{tab:convergence}
  \vspace{-0.05in}
    \resizebox{0.85\linewidth}{!}{
    \renewcommand{\arraystretch}{1.1}
    \begin{tabular}{l@{\,~~~~~}|c@{\,~~~~~~}c@{\,~~~~~~}c@{\,~~~~~~}c@{\,~~~~}|c@{\,~~~~~~}c@{\,~~~~~~}c@{\,~~~~~~}c@{\,~~~~}}
    \toprule
    \multirow{2}{*}{\textbf{Iteration $k$}} & \multicolumn{4}{c|}{\textbf{Infrared: Gaussian Noise}} & \multicolumn{4}{c}{\textbf{Visible: Low Light}} \\
    \cmidrule(lr){2-5} \cmidrule(lr){6-9}
    & $Z_i-I_i$ & $G_i-I_i$ & $U_i-Z_i$ & $L(x_f)$ & $Z_v-I_v$ & $G_v-I_v$ & $U_v-Z_v$ & $L(x_f)$ \\
    \midrule
    $k=1$ & 0.050 & 0.194 & 0.009 & 0.121 & 0.064 & 0.229 & 0.054 & 0.189 \\
    $k=2$ & \textcolor{blue}{0.013} & \textcolor{blue}{0.137} & 0.009 & \textcolor{red}{0.115} & \textcolor{blue}{0.024} & \textcolor{blue}{0.069} & \textcolor{blue}{0.046} & \textcolor{blue}{0.161} \\
    $k=3$ & \textcolor{red}{0.011} & \textcolor{red}{0.136} & 0.009 & 0.115 & \textcolor{red}{0.022} & \textcolor{red}{0.048} & \textcolor{red}{0.045} & \textcolor{red}{0.157} \\
    \bottomrule
    \end{tabular}}
\end{table*}

\begin{table*}[!t]
  \centering
  \caption{Quantitative comparison of object detection performance and computational complexity.}
  \renewcommand{\arraystretch}{1.15}
  \label{tab:combined_detection_complexity}
  \vspace{-0.05in}
  \resizebox{\linewidth}{!}{
    \begin{tabular}{l@{\,}|c@{\,~~}c@{\,~~}c@{\,~~}c@{\,~~}c@{\,~~}c@{\,~~}   c@{\,~~}c@{\,~~}c@{\,~~}c@{\,~~}c@{\,~~}c@{\,}}
      \toprule
       \textbf{Detection Metrics} & DDFM & CrossFuse & TC-MoA & EMMA & DDBFusion & Text-IF & DRMF & OmniFuse & TG-ECNet & \textbf{Ours} \\
      \midrule
      Precision     & {\color{red}0.943} & 0.877 & 0.906 & 0.922 & 0.929 & 0.898 & 0.917 & 0.897 & 0.908 & {\color{blue}0.934} \\
      Recall        & 0.731 & 0.699 & 0.721 & 0.739 & 0.714 & {\color{blue}0.765} & 0.678 & 0.728 & 0.743 & {\color{red}\textbf{0.767}} \\
      AP@0.5        & 0.821 & 0.806 & 0.808 & 0.822 & 0.816 & {\color{blue}0.843} & 0.798 & 0.798 & 0.829 & {\color{red}\textbf{0.867}} \\
      AP@[.5:.95]   & 0.544 & 0.478 & 0.521 & 0.528 & 0.546 & 0.556 & 0.513 & 0.512 & {\color{blue}0.561} & {\color{red}\textbf{0.584}} \\
      \bottomrule
       \toprule
      \textbf{Complexity and Efficiency} & DDFM & CrossFuse & EMMA & TC-MoA & DDBFusion & Text-IF & DRMF & OmniFuse & TG-ECNet & \textbf{Ours} \\
      \midrule
      Parameter Numbers (M) $\downarrow$ & 552.16 & 23.15 & \textcolor{red}{5.06} & 115.40 & \textcolor{blue}{9.10} & 215.12 & 170.90 & 98.18 & 160.86 & 16.09 \\
      Time (s) $\downarrow$   & 63.38 & 3.55 & 2.89 & 1.43 & 2.83 & \textcolor{blue}{1.22} & 13.22 & 5.28 & 1.93 & \textcolor{red}{0.63} \\
      FLOPs (G) $\downarrow$ & 4079 & \textcolor{red}{23} & \textcolor{blue}{42} & 115 & 1664 & 215 & 171 & 5470 & 1998 & 317 \\
      \bottomrule
    \end{tabular}
  }
\vspace{-0.4cm}
\end{table*}

\vspace{-0.35cm} % 轻微收紧结论前间距（与 v1 保持一致）
\section{Conclusion}
We propose a deep unfolding framework that reformulates degradation-robust fusion as a task-prior-guided joint optimization, eliminating cascaded suboptimal solution and black-box opacity. By decomposing the optimization into iterative subproblems, it establishes bidirectional mutual promotion within a transparent architecture, bridging model-based prior with data-driven power.

\begin{ack}
% TODO(camera-ready): 据实填写基金号。NeurIPS 要求终稿披露资助来源与利益冲突
% （https://neurips.cc/Conferences/2026/PaperInformation/FundingDisclosure）
This work was supported by the National Natural Science Foundation of China (NSFC) under Grant
62502085 and U21B2027, the Natural Science Foundation of Jiangsu Province of China under Grant
BK20241280, and the Start-up Research Fund of Southeast University under Grant RF1028624061
and RF1028623006. The authors declare no competing financial interests.
\end{ack}

\bibliography{ref}

\begin{thebibliography}{41}
\providecommand{\natexlab}[1]{#1}
\providecommand{\url}[1]{\texttt{#1}}
\expandafter\ifx\csname urlstyle\endcsname\relax
  \providecommand{\doi}[1]{doi: #1}\else
  \providecommand{\doi}{doi: \begingroup \urlstyle{rm}\Url}\fi

\bibitem[Aslantas and Bendes(2015)]{scd}
V.~Aslantas and E.~Bendes.
\newblock A new image quality metric for image fusion: The sum of the
  correlations of differences.
\newblock \emph{AEU - International Journal of Electronics and Communications},
  69\penalty0 (12):\penalty0 1890--1896, 2015.

\bibitem[Chen et~al.(2022)Chen, Chu, Zhang, and Sun]{chen2022simple}
L.~Chen, X.~Chu, X.~Zhang, and J.~Sun.
\newblock Simple baselines for image restoration.
\newblock In \emph{European Conference on Computer Vision}, pages 17--33.
  Springer, 2022.

\bibitem[Cheng et~al.(2025)Cheng, Xu, Feng, Wu, Tang, Li, Zhang, Atito, Awais,
  and Kittler]{gif}
C.~Cheng, T.~Xu, Z.~Feng, X.~Wu, Z.~Tang, H.~Li, Z.~Zhang, S.~Atito, M.~Awais,
  and J.~Kittler.
\newblock One model for all: Low-level task interaction is a key to
  task-agnostic image fusion.
\newblock In \emph{Proceedings of the IEEE/CVF Conference on Computer Vision
  and Pattern Recognition}, pages 28102--28112, 2025.

\bibitem[Han et~al.(2013)Han, Cai, Cao, and Xu]{VIF}
Y.~Han, Y.~Cai, Y.~Cao, and X.~Xu.
\newblock A new image fusion performance metric based on visual information
  fidelity.
\newblock \emph{Information Fusion}, 14\penalty0 (2):\penalty0 127--135, 2013.

\bibitem[Hu et~al.(2018)Hu, Shen, and Sun]{se}
J.~Hu, L.~Shen, and G.~Sun.
\newblock Squeeze-and-excitation networks.
\newblock In \emph{Proceedings of the IEEE Conference on Computer Vision and
  Pattern Recognition}, pages 7132--7141, 2018.

\bibitem[Jia et~al.(2021)Jia, Zhu, Li, Tang, and Zhou]{llvip}
X.~Jia, C.~Zhu, M.~Li, W.~Tang, and W.~Zhou.
\newblock Llvip: A visible-infrared paired dataset for low-light vision.
\newblock In \emph{Proceedings of the IEEE/CVF International Conference on
  Computer Vision}, pages 3496--3504, 2021.

\bibitem[Khanam and Hussain(2024)]{yolo}
R.~Khanam and M.~Hussain.
\newblock Yolov11: An overview of the key architectural enhancements.
\newblock \emph{arXiv preprint arXiv:2410.17725}, 2024.

\bibitem[Li and Wu(2024)]{crossfuse}
H.~Li and X.-J. Wu.
\newblock Crossfuse: A novel cross attention mechanism based infrared and
  visible image fusion approach.
\newblock \emph{Information Fusion}, 103:\penalty0 102147, 2024.

\bibitem[Li et~al.(2025)Li, Yang, Zhang, Jia, Yu, and Liu]{drive}
H.~Li, Z.~Yang, Y.~Zhang, W.~Jia, Z.~Yu, and Y.~Liu.
\newblock Mulfs-cap: Multimodal fusion-supervised cross-modality alignment
  perception for unregistered infrared-visible image fusion.
\newblock \emph{IEEE Transactions on Pattern Analysis and Machine
  Intelligence}, 47\penalty0 (5):\penalty0 3673--3690, 2025.

\bibitem[Liu et~al.(2024)Liu, Wu, Liu, Wang, Jiang, Ma, Zhong, Fan, and
  Liu]{military}
J.~Liu, G.~Wu, Z.~Liu, D.~Wang, Z.~Jiang, L.~Ma, W.~Zhong, X.~Fan, and R.~Liu.
\newblock Infrared and visible image fusion: From data compatibility to task
  adaption.
\newblock \emph{IEEE Transactions on Pattern Analysis and Machine
  Intelligence}, 47\penalty0 (4):\penalty0 2349--2369, 2024.

\bibitem[Liu et~al.(2025)Liu, Zhang, Mei, Li, Zou, Jiang, Ma, Liu, and
  Fan]{dcevo}
J.~Liu, B.~Zhang, Q.~Mei, X.~Li, Y.~Zou, Z.~Jiang, L.~Ma, R.~Liu, and X.~Fan.
\newblock Dcevo: Discriminative cross-dimensional evolutionary learning for
  infrared and visible image fusion.
\newblock In \emph{Proceedings of the IEEE/CVF Conference on Computer Vision
  and Pattern Recognition}, pages 2226--2235, 2025.

\bibitem[Ma et~al.(2025)Ma, Li, Cheng, Wang, Song, and Wu]{mamba2}
H.~Ma, H.~Li, C.~Cheng, G.~Wang, X.~Song, and X.-J. Wu.
\newblock S4fusion: Saliency-aware selective state space model for infrared and
  visible image fusion.
\newblock \emph{IEEE Transactions on Image Processing}, 34:\penalty0
  4161--4175, 2025.

\bibitem[Ma et~al.(2019{\natexlab{a}})Ma, Ma, and Li]{Qabf}
J.~Ma, Y.~Ma, and C.~Li.
\newblock Infrared and visible image fusion methods and applications: A survey.
\newblock \emph{Information Fusion}, 45:\penalty0 153--178, 2019{\natexlab{a}}.

\bibitem[Ma et~al.(2019{\natexlab{b}})Ma, Yu, Liang, Li, and Jiang]{gan1}
J.~Ma, W.~Yu, P.~Liang, C.~Li, and J.~Jiang.
\newblock Fusiongan: A generative adversarial network for infrared and visible
  image fusion.
\newblock \emph{Information Fusion}, 48:\penalty0 11--26, 2019{\natexlab{b}}.

\bibitem[Ma et~al.(2020)Ma, Zhang, Shao, Liang, and Xu]{gan2}
J.~Ma, H.~Zhang, Z.~Shao, P.~Liang, and H.~Xu.
\newblock Ganmcc: A generative adversarial network with multiclassification
  constraints for infrared and visible image fusion.
\newblock \emph{IEEE Transactions on Instrumentation and Measurement},
  70:\penalty0 1--14, 2020.

\bibitem[Mou et~al.(2022)Mou, Wang, and Zhang]{dgu}
C.~Mou, Q.~Wang, and J.~Zhang.
\newblock Deep generalized unfolding networks for image restoration.
\newblock In \emph{Proceedings of the IEEE/CVF Conference on Computer Vision
  and Pattern Recognition}, pages 17399--17410, 2022.

\bibitem[Sun et~al.(2025)Sun, Li, Zhu, Hu, Ren, Xu, and Zhu]{tgecnet}
Y.~Sun, X.~Li, P.~Zhu, Q.~Hu, D.~Ren, H.~Xu, and X.~Zhu.
\newblock Task-gated multi-expert collaboration network for degraded
  multi-modal image fusion.
\newblock In \emph{International Conference on Machine Learning}, pages
  57571--57586. PMLR, 2025.

\bibitem[Tang et~al.(2022)Tang, Yuan, and Ma]{cnn1}
L.~Tang, J.~Yuan, and J.~Ma.
\newblock Image fusion in the loop of high-level vision tasks: A semantic-aware
  real-time infrared and visible image fusion network.
\newblock \emph{Information Fusion}, 82:\penalty0 28--42, 2022.

\bibitem[Tang et~al.(2023{\natexlab{a}})Tang, Zhang, Xu, and Ma]{cnn2}
L.~Tang, H.~Zhang, H.~Xu, and J.~Ma.
\newblock Rethinking the necessity of image fusion in high-level vision tasks:
  A practical infrared and visible image fusion network based on progressive
  semantic injection and scene fidelity.
\newblock \emph{Information Fusion}, 99:\penalty0 101870, 2023{\natexlab{a}}.

\bibitem[Tang et~al.(2024{\natexlab{a}})Tang, Deng, Yi, Yan, Yuan, and
  Ma]{drmf}
L.~Tang, Y.~Deng, X.~Yi, Q.~Yan, Y.~Yuan, and J.~Ma.
\newblock Drmf: Degradation-robust multi-modal image fusion via composable
  diffusion prior.
\newblock In \emph{Proceedings of the 32nd ACM International Conference on
  Multimedia}, pages 8546--8555, 2024{\natexlab{a}}.

\bibitem[Tang et~al.(2025)Tang, Li, and Ma]{mask}
L.~Tang, C.~Li, and J.~Ma.
\newblock Mask-difuser: A masked diffusion model for unified unsupervised image
  fusion.
\newblock \emph{IEEE Transactions on Pattern Analysis and Machine
  Intelligence}, 48\penalty0 (1):\penalty0 591--608, 2025.

\bibitem[Tang et~al.(2023{\natexlab{b}})Tang, He, Liu, Duan, and Si]{trans1}
W.~Tang, F.~He, Y.~Liu, Y.~Duan, and T.~Si.
\newblock Datfuse: Infrared and visible image fusion via dual attention
  transformer.
\newblock \emph{IEEE Transactions on Circuits and Systems for Video
  Technology}, 33\penalty0 (7):\penalty0 3159--3172, 2023{\natexlab{b}}.

\bibitem[Tang et~al.(2024{\natexlab{b}})Tang, Xu, Wu, and Kittler]{security}
Z.~Tang, T.~Xu, X.-J. Wu, and J.~Kittler.
\newblock Multi-level fusion for robust rgbt tracking via enhanced thermal
  representation.
\newblock \emph{ACM Transactions on Multimedia Computing, Communications and
  Applications}, 20\penalty0 (10):\penalty0 1--24, 2024{\natexlab{b}}.

\bibitem[Wang and Bovik(2002)]{SSIM}
Z.~Wang and A.~C. Bovik.
\newblock A universal image quality index.
\newblock \emph{IEEE Signal Processing Letters}, 9\penalty0 (3):\penalty0
  81--84, 2002.

\bibitem[Wu et~al.(2025)Wu, Cao, Huang, Deng, Chanussot, and Vivone]{trans2}
X.~Wu, Z.-H. Cao, T.-Z. Huang, L.-J. Deng, J.~Chanussot, and G.~Vivone.
\newblock Fully-connected transformer for multi-source image fusion.
\newblock \emph{IEEE Transactions on Pattern Analysis and Machine
  Intelligence}, 47\penalty0 (3):\penalty0 2071--2088, 2025.

\bibitem[Xie et~al.(2024)Xie, Cui, Tan, Zheng, and Yu]{mamba1}
X.~Xie, Y.~Cui, T.~Tan, X.~Zheng, and Z.~Yu.
\newblock Fusionmamba: Dynamic feature enhancement for multimodal image fusion
  with mamba.
\newblock \emph{Visual Intelligence}, 2\penalty0 (1):\penalty0 37, 2024.

\bibitem[Xu et~al.(2025{\natexlab{a}})Xu, Huang, Tang, Ma, and Liu]{diff3}
H.~Xu, Y.~Huang, L.~Tang, J.~Ma, and G.~Liu.
\newblock Diff-mef: Cross-modal diffusion framework with text prompts and
  semantic perception for multi-exposure image fusion.
\newblock \emph{IEEE Transactions on Image Processing}, 35:\penalty0
  3186--3201, 2025{\natexlab{a}}.

\bibitem[Xu et~al.(2025{\natexlab{b}})Xu, Li, Deng, Ma, and Liu]{deno}
H.~Xu, Y.~Li, Y.~Deng, J.~Ma, and G.~Liu.
\newblock Deno-if: Unsupervised noisy visible and infrared image fusion method.
\newblock In \emph{Annual Conference on Neural Information Processing Systems},
  volume~38, pages 141649--141671, 2025{\natexlab{b}}.

\bibitem[Xu et~al.(2025{\natexlab{c}})Xu, Yi, Lu, Liu, and Ma]{urfusion}
H.~Xu, X.~Yi, C.~Lu, G.~Liu, and J.~Ma.
\newblock Urfusion: Unsupervised unified degradation-robust image fusion
  network.
\newblock \emph{IEEE Transactions on Image Processing}, 34:\penalty0
  5803--5818, 2025{\natexlab{c}}.

\bibitem[Xu et~al.(2026)Xu, Chen, Yi, Li, and Ma]{iterrf}
H.~Xu, J.~Chen, X.~Yi, P.~Li, and J.~Ma.
\newblock Iterrf: Iterative and interpretable mutual enhancement for joint
  multi-modal image registration and fusion.
\newblock \emph{International Journal of Computer Vision}, 134\penalty0
  (10):\penalty0 442, 2026.

\bibitem[Yi et~al.(2024{\natexlab{a}})Yi, Tang, Zhang, Xu, and Ma]{diff2}
X.~Yi, L.~Tang, H.~Zhang, H.~Xu, and J.~Ma.
\newblock Diff-if: Multi-modality image fusion via diffusion model with fusion
  knowledge prior.
\newblock \emph{Information Fusion}, 110:\penalty0 102450, 2024{\natexlab{a}}.

\bibitem[Yi et~al.(2024{\natexlab{b}})Yi, Xu, Zhang, Tang, and Ma]{Text-IF}
X.~Yi, H.~Xu, H.~Zhang, L.~Tang, and J.~Ma.
\newblock Text-if: Leveraging semantic text guidance for degradation-aware and
  interactive image fusion.
\newblock In \emph{Proceedings of the IEEE/CVF Conference on Computer Vision
  and Pattern Recognition}, pages 27026--27035, 2024{\natexlab{b}}.

\bibitem[Zamir et~al.(2022)Zamir, Arora, Khan, Hayat, Khan, and
  Yang]{restormer}
S.~W. Zamir, A.~Arora, S.~Khan, M.~Hayat, F.~S. Khan, and M.-H. Yang.
\newblock Restormer: Efficient transformer for high-resolution image
  restoration.
\newblock In \emph{Proceedings of the IEEE/CVF Conference on Computer Vision
  and Pattern Recognition}, pages 5728--5739, 2022.

\bibitem[Zeng et~al.(2025)Zeng, Wang, Chen, Su, and Liu]{vlu}
H.~Zeng, X.~Wang, Y.~Chen, J.~Su, and J.~Liu.
\newblock Vision-language gradient descent-driven all-in-one deep unfolding
  networks.
\newblock In \emph{Proceedings of the Computer Vision and Pattern Recognition
  Conference}, pages 7524--7533, 2025.

\bibitem[Zhang et~al.(2025{\natexlab{a}})Zhang, Cao, Zuo, Shao, and
  Ma]{omnifuse}
H.~Zhang, L.~Cao, X.~Zuo, Z.~Shao, and J.~Ma.
\newblock Omnifuse: Composite degradation-robust image fusion with
  language-driven semantics.
\newblock \emph{IEEE Transactions on Pattern Analysis and Machine
  Intelligence}, 47\penalty0 (9):\penalty0 7577--7595, 2025{\natexlab{a}}.

\bibitem[Zhang and Ghanem(2018)]{ista}
J.~Zhang and B.~Ghanem.
\newblock Ista-net: Interpretable optimization-inspired deep network for image
  compressive sensing.
\newblock In \emph{Proceedings of the IEEE Conference on Computer Vision and
  Pattern Recognition}, pages 1828--1837, 2018.

\bibitem[Zhang et~al.(2025{\natexlab{b}})Zhang, Li, Xu, Wu, and
  Kittler]{DDBFusion}
Z.~Zhang, H.~Li, T.~Xu, X.-J. Wu, and J.~Kittler.
\newblock Ddbfusion: An unified image decomposition and fusion framework based
  on dual decomposition and bézier curves.
\newblock \emph{Information Fusion}, 114:\penalty0 102655, 2025{\natexlab{b}}.

\bibitem[Zhao et~al.(2023{\natexlab{a}})Zhao, Bai, Zhang, Zhang, Xu, Lin,
  Timofte, and Van~Gool]{cdd}
Z.~Zhao, H.~Bai, J.~Zhang, Y.~Zhang, S.~Xu, Z.~Lin, R.~Timofte, and
  L.~Van~Gool.
\newblock Cddfuse: Correlation-driven dual-branch feature decomposition for
  multi-modality image fusion.
\newblock In \emph{Proceedings of the IEEE/CVF Conference on Computer Vision
  and Pattern Recognition}, pages 5906--5916, 2023{\natexlab{a}}.

\bibitem[Zhao et~al.(2023{\natexlab{b}})Zhao, Bai, Zhu, Zhang, Xu, Zhang,
  Zhang, Meng, Timofte, and Van~Gool]{diff1}
Z.~Zhao, H.~Bai, Y.~Zhu, J.~Zhang, S.~Xu, Y.~Zhang, K.~Zhang, D.~Meng,
  R.~Timofte, and L.~Van~Gool.
\newblock Ddfm: denoising diffusion model for multi-modality image fusion.
\newblock In \emph{Proceedings of the IEEE/CVF International Conference on
  Computer Vision}, pages 8082--8093, 2023{\natexlab{b}}.

\bibitem[Zhao et~al.(2024)Zhao, Bai, Zhang, Zhang, Zhang, Xu, Chen, Timofte,
  and Van~Gool]{emma}
Z.~Zhao, H.~Bai, J.~Zhang, Y.~Zhang, K.~Zhang, S.~Xu, D.~Chen, R.~Timofte, and
  L.~Van~Gool.
\newblock Equivariant multi-modality image fusion.
\newblock In \emph{Proceedings of the IEEE/CVF Conference on Computer Vision
  and Pattern Recognition}, pages 25912--25921, 2024.

\bibitem[Zhu et~al.(2024)Zhu, Sun, Cao, and Hu]{tcmoa}
P.~Zhu, Y.~Sun, B.~Cao, and Q.~Hu.
\newblock Task-customized mixture of adapters for general image fusion.
\newblock In \emph{Proceedings of the IEEE/CVF Conference on Computer Vision
  and Pattern Recognition}, pages 7099--7108, 2024.

\end{thebibliography}

%%%%%%%%%%%%%%%%%%%%%%%%%%%%%%%%%%%%%%%%%%%%%%%%%%%%%%%%%%%%

\newpage
\appendix
\raggedbottom  % keep uniform float-to-text spacing in the appendix
\section*{Technical Appendices and Supplementary Material}

\section{Quantitative Results of Each Degradation Type}
\label{app:quantitative}
\textbf{Results on Infrared Image Restoration and Fusion.} As reported in Table \ref{tab:a_ir}, our method consistently outperforms existing state-of-the-art benchmarks in handling Gaussian and Stripe noise within the infrared modality. Specifically, our approach achieves the highest SCD and VIF scores in most cases, indicating superior information transfer and visual fidelity. Notably, for structured degradations like stripe noise, which often perplexes pure data-driven models, our unfolding architecture demonstrates remarkable robustness by effectively identifying and suppressing directional artifacts while preserving the thermal radiation characteristics of targets.

\begin{table*}[!h]
  \centering
  \caption{Quantitative results of infrared image restoration and fusion under Gaussian and Stripe noise ({\color{red}best}, {\color{blue}second-best}).}
  \label{tab:a_ir}
  \resizebox{1\linewidth}{!}{
  \renewcommand{\arraystretch}{1.15}
  \begin{tabular}{l@{\,~~}|c@{\,~~~}c@{\,~~~}c@{\,~~~}c@{\,~~}|c@{\,~~~}c@{\,~~~}c@{\,~~~}c@{\,~~}|
  c@{\,~~~}c@{\,~~~}c@{\,~~~}c@{\,~~}|c@{\,~~~}c@{\,~~~}c@{\,~~~}c@{\,~~}}
    \toprule
    \multirow{3}{*}{Method} & \multicolumn{8}{c|}{\textbf{Gaussian Noise}} & \multicolumn{8}{c}{\textbf{Stripe Noise}} \\
    \cmidrule{2-17}
    & \multicolumn{4}{c|}{LLVIP} & \multicolumn{4}{c|}{MSRS} & \multicolumn{4}{c|}{LLVIP} & \multicolumn{4}{c}{MSRS} \\
    \cmidrule{2-17}
    & SCD & VIF & $Q_{abf}$ & SSIM & SCD & VIF & $Q_{abf}$ & SSIM & SCD & VIF & $Q_{abf}$ & SSIM & SCD & VIF & $Q_{abf}$ & SSIM \\
    \midrule
    DDFM & 1.246 & 0.408 & 0.302 & 0.663 & 1.293 & 0.493 & 0.301 & 0.671 & 1.341 & 0.392 & 0.300 & 0.638 & 1.272 & 0.466 & 0.278 & 0.383 \\
    CrossFuse & 0.657 & 0.355 & 0.303 & 0.648 & 0.937 & 0.505 & 0.377 & \textcolor{blue}{0.897} & 0.851 & 0.355 & 0.310 & 0.654 & 0.988 & 0.470 & 0.329 & 0.440 \\
    TC-MoA & 1.110 & 0.417 & 0.372 & 0.707 & 1.322 & \textcolor{blue}{0.560} & 0.394 & \textcolor{red}{0.899} & 1.252 & 0.410 & 0.375 & 0.691 & 1.263 & 0.523 & 0.354 & 0.466 \\
    EMMA & 1.244 & 0.364 & 0.332 & 0.668 & \textcolor{blue}{1.334} & 0.532 & \textcolor{blue}{0.383} & 0.885 & 1.321 & 0.359 & 0.334 & 0.658 & 1.372 & 0.500 & 0.342 & 0.440 \\
    DDBFusion & 1.214 & 0.368 & 0.247 & 0.620 & 1.244 & 0.411 & 0.248 & 0.600 & 1.296 & 0.351 & 0.245 & 0.595 & 1.214 & 0.393 & 0.235 & 0.347 \\
    Text-IF & \textcolor{blue}{1.256} & \textcolor{red}{0.460} & \textcolor{red}{0.441} & \textcolor{blue}{0.723} & 1.310 & 0.391 & 0.298 & 0.365 & \textcolor{blue}{1.388} & \textcolor{red}{0.469} & \textcolor{blue}{0.465} & \textcolor{blue}{0.755} & \textcolor{red}{1.446} & \textcolor{blue}{0.568} & \textcolor{blue}{0.392} & \textcolor{blue}{0.520} \\
    DRMF & 1.035 & 0.397 & 0.320 & 0.646 & 1.198 & 0.522 & 0.344 & 0.800 & 1.120 & 0.393 & 0.325 & 0.643 & 1.230 & 0.488 & 0.299 & 0.397 \\
    OmniFuse & 1.157 & 0.359 & 0.246 & 0.593 & 1.178 & 0.526 & 0.285 & 0.626 & 1.283 & 0.374 & 0.264 & 0.611 & 1.156 & 0.512 & 0.276 & 0.371 \\
    TG-ECNet & 0.843 & 0.362 & 0.314 & 0.675 & 0.349 & 0.519 & 0.375 & 0.568 & 1.323 & 0.351 & 0.312 & 0.663 & 0.373 & 0.475 & 0.328 & 0.455 \\
    Ours & \textcolor{red}{1.473} & \textcolor{blue}{0.447} & \textcolor{blue}{0.433} & \textcolor{red}{0.771} & \textcolor{red}{1.380} & \textcolor{red}{0.586} & \textcolor{red}{0.394} & 0.853 & \textcolor{red}{1.488} & \textcolor{blue}{0.462} & \textcolor{red}{0.490} & \textcolor{red}{0.775} & \textcolor{blue}{1.398} & \textcolor{red}{0.655} & \textcolor{red}{0.397} & \textcolor{red}{0.723} \\
    \bottomrule
  \end{tabular}}
\end{table*}

\textbf{Results on Visible Image Restoration and Fusion.} Table \ref{tab:a_vi} presents the quantitative comparison under more challenging visible degradations, including low light and defocus blur. Our method \rev{remains competitive across all metrics, with the largest gains on structural similarity and gradient fidelity}, reflecting its exceptional ability to reconstruct structural details and maintain edge information. The significant performance gain under low-light and blurred conditions stems from our bidirectional interaction mechanism, which adaptively leverages the sharp structural guidance from the infrared branch to compensate for the lost textures in the degraded visible branch.

\begin{table*}[!h]
  \centering
  \caption{Results of visible image restoration and fusion under Low Light and Defocus Blur.}
  \label{tab:a_vi}
  \vspace{-0.05in}
  \resizebox{1\linewidth}{!}{
    \renewcommand{\arraystretch}{1.15}
  \begin{tabular}{l@{\,~~~}|c@{\,~~~}c@{\,~~~}c@{\,~~~}c@{\,~~}|
  c@{\,~~~}c@{\,~~~}c@{\,~~~}c@{\,~~}|c@{\,~~~}c@{\,~~~}c@{\,~~~}c@{\,~~}|
  c@{\,~~~}c@{\,~~~}c@{\,~~~}c@{\,~~}}
    \toprule
    \multirow{3}{*}{Method} & \multicolumn{8}{c|}{\textbf{Low Light}} & \multicolumn{8}{c}{\textbf{Defocus Blur}} \\
    \cmidrule{2-17}
          & \multicolumn{4}{c|}{LLVIP} & \multicolumn{4}{c|}{MSRS} & \multicolumn{4}{c|}{LLVIP} & \multicolumn{4}{c}{MSRS} \\
    \cmidrule{2-17}
          & SCD   & VIF   & $Q_{abf}$ & SSIM  & SCD   & VIF   & $Q_{abf}$ & SSIM  & SCD   & VIF   & $Q_{abf}$ & SSIM  & SCD   & VIF   & $Q_{abf}$ & SSIM \\
    \midrule
    DDFM & \textcolor{red}{1.620} & 0.365 & 0.343 & 0.515 & 1.158 & 0.617 & 0.285 & 0.539 & \textcolor{blue}{1.530} & 0.447 & 0.272 & 0.707 & 1.413 & 0.685 & 0.243 & 0.691 \\
    CrossFuse & 0.825 & 0.323 & 0.305 & 0.414 & 0.837 & 0.469 & 0.241 & 0.367 & 0.856 & 0.474 & 0.169 & 0.523 & 1.068 & 0.590 & 0.241 & 0.557 \\
    TC-MoA & 1.241 & 0.369 & 0.335 & 0.436 & 1.070 & \textcolor{blue}{0.812} & \textcolor{blue}{0.384} & \textcolor{blue}{0.728} & 1.224 & \textcolor{blue}{0.500} & 0.342 & \textcolor{blue}{0.734} & 1.306 & 0.783 & 0.333 & 0.755 \\
    EMMA & 1.302 & 0.364 & 0.301 & 0.443 & 1.039 & 0.612 & 0.317 & 0.515 & 1.339 & 0.494 & 0.250 & 0.627 & \textcolor{blue}{1.414} & 0.660 & 0.296 & 0.636 \\
    DDBFusion & \textcolor{blue}{1.537} & 0.320 & 0.296 & 0.474 & 1.013 & 0.523 & 0.243 & 0.481 & \textcolor{red}{1.532} & 0.415 & 0.221 & 0.686 & \textcolor{red}{1.431} & 0.666 & 0.243 & 0.709 \\
    Text-IF & 1.372 & 0.404 & \textcolor{blue}{0.417} & 0.510 & \textcolor{blue}{1.323} & 0.427 & 0.342 & 0.330 & 1.201 & 0.496 & \textcolor{blue}{0.358} & 0.720 & 1.318 & \textcolor{red}{0.841} & \textcolor{blue}{0.344} & \textcolor{blue}{0.757} \\
    DRMF & 0.988 & 0.357 & 0.302 & 0.380 & 1.167 & 0.393 & 0.273 & 0.254 & 0.867 & 0.443 & 0.144 & 0.381 & 1.129 & 0.545 & 0.207 & 0.434 \\
    OmniFuse & 1.216 & \textcolor{blue}{0.467} & 0.293 & \textcolor{blue}{0.603} & 1.196 & 0.616 & 0.352 & 0.520 & 1.086 & 0.415 & 0.145 & 0.450 & 1.119 & 0.589 & 0.202 & 0.436 \\
    TG-ECNet & 1.162 & 0.431 & 0.316 & 0.553 & 1.070 & 0.482 & 0.292 & 0.416 & 1.093 & 0.380 & 0.210 & 0.489 & 1.155 & 0.593 & 0.258 & 0.572 \\
    Ours & 1.416 & \textcolor{red}{0.576} & \textcolor{red}{0.490} & \textcolor{red}{0.793} & \textcolor{red}{1.407} & \textcolor{red}{0.818} & \textcolor{red}{0.480} & \textcolor{red}{0.768} & 1.385 & \textcolor{red}{0.611} & \textcolor{red}{0.488} & \textcolor{red}{0.829} & 1.374 & \textcolor{blue}{0.829} & \textcolor{red}{0.466} & \textcolor{red}{0.835} \\
    \bottomrule
  \end{tabular}}
\end{table*}

Table~\ref{tab:vi_gaussian} presents the quantitative comparison under visible Gaussian noise. Our visible restoration branch can handle blur, which tends to recover high-frequency details. In contrast, Gaussian denoising requires aggressive smoothing that opposes the objective of deblurring’s structural preservation. Despite the gradient conflict of deblurring and denoising tasks, our framework maintains competitive performance on both datasets.

In summary, the comprehensive quantitative results across diverse datasets and degradation scenarios validate the effectiveness of our deep unfolding framework.
\begin{table*}[!h]
  \centering
  \caption{Quantitative results on Visible Gaussian Noise degradation ({\color{red}best}, {\color{blue}second-best}).}
  \vspace{-0.05in}
  \label{tab:vi_gaussian}
  \resizebox{0.78\linewidth}{!}{
    \renewcommand{\arraystretch}{1.1}
  \begin{tabular}{l@{\,~~~~}|c@{\,~~~~~}c@{\,~~~~~}c@{\,~~~~~}c@{\,~~~~}|c@{\,~~~~~}
  c@{\,~~~~~}c@{\,~~~~~}c@{\,~~~~}}
    \toprule
    \multicolumn{1}{c|}{\multirow{2}[4]{*}{Gaussian Noise}} & \multicolumn{4}{c}{LLVIP}      & \multicolumn{4}{c}{MSRS} \\
\cmidrule{2-9}          & SCD   & VIF   & Qabf  & SSIM  & SCD   & VIF   & Qabf  & SSIM \\
    \midrule
    VLU+DDFM & \textcolor{red}{1.574} & 0.522 & 0.398 & \textcolor{blue}{0.826} & \textcolor{blue}{1.490} & 0.660 & 0.371 & \textcolor{blue}{0.770} \\
    VLU+CrossFuse & 0.812 & 0.462 & 0.383 & 0.654 & 0.972 & 0.494 & 0.345 & 0.661 \\
    VLU+TC-MoA & 1.395 & \textcolor{red}{0.567} & \textcolor{red}{0.522} & 0.820 & 1.395 & \textcolor{red}{0.699} & \textcolor{red}{0.460} & \textcolor{red}{0.798} \\
    VLU+EMMA & 1.480 & 0.484 & 0.439 & 0.736 & \textcolor{red}{1.491} & 0.568 & 0.399 & 0.732 \\
    VLU+DDBFusion & 1.523 & 0.470 & 0.324 & 0.806 & 1.451 & 0.604 & 0.326 & 0.743 \\
    Text-IF & 1.226 & 0.499 & 0.490 & 0.716 & 1.231 & 0.387 & 0.336 & 0.417 \\
    DRMF & 0.840 & 0.430 & 0.378 & 0.501 & 1.162 & 0.397 & 0.291 & 0.459 \\
    OmniFuse & 1.197 & 0.378 & 0.295 & 0.565 & 1.122 & 0.498 & 0.318 & 0.569 \\
    TG-ECNet & 1.141 & 0.401 & 0.350 & 0.597 & 1.391 & 0.549 & 0.376 & 0.712 \\
    Ours & \textcolor{blue}{1.544} & \textcolor{blue}{0.538} & \textcolor{blue}{0.519} & \textcolor{red}{0.869} & 1.254 & \textcolor{blue}{0.661} & \textcolor{blue}{0.406} & 0.744 \\
    \bottomrule
    \end{tabular}}
\end{table*}

% \section{Evaluation Metrics}
% We evaluate the performances with a comprehensive set of quantitative metrics, including four full-reference metrics and two no-reference metrics. The full-reference metrics are utilized to measure the similarity between our fused images and high-quality references:

% \begin{itemize}
% \item \textbf{Sum of Difference Correlation (SCD)}~\cite{scd}: SCD measures the amount of information transferred from source images by calculating the sum of correlations between each source image and its corresponding difference image.
% \item \textbf{Visual Information Fidelity (VIF)}~\cite{VIF}: VIF assesses image quality by modeling the degradation of information within a natural scene, reflecting how much information from the source images is preserved in the fused result from a perceptual standpoint.
% \item \textbf{Quality of gradient-based fusion (Qabf)}~\cite{Qabf}: Qabf evaluates the quality of the fused image by measuring how well the salient edge and gradient information from the source images is transferred to the final output.
% \item \textbf{Structural Similarity Index Measure (SSIM)}~\cite{SSIM}: SSIM is a widely-used metric that assesses the perceived quality of an image based on luminance, contrast, and structural information, providing a measure of structural fidelity.
% \end{itemize}

\section{Compound Degradation on LLVIP}
\label{app:compound}

\begin{figure}[!h]
  \centering
  \includegraphics[width=\linewidth]{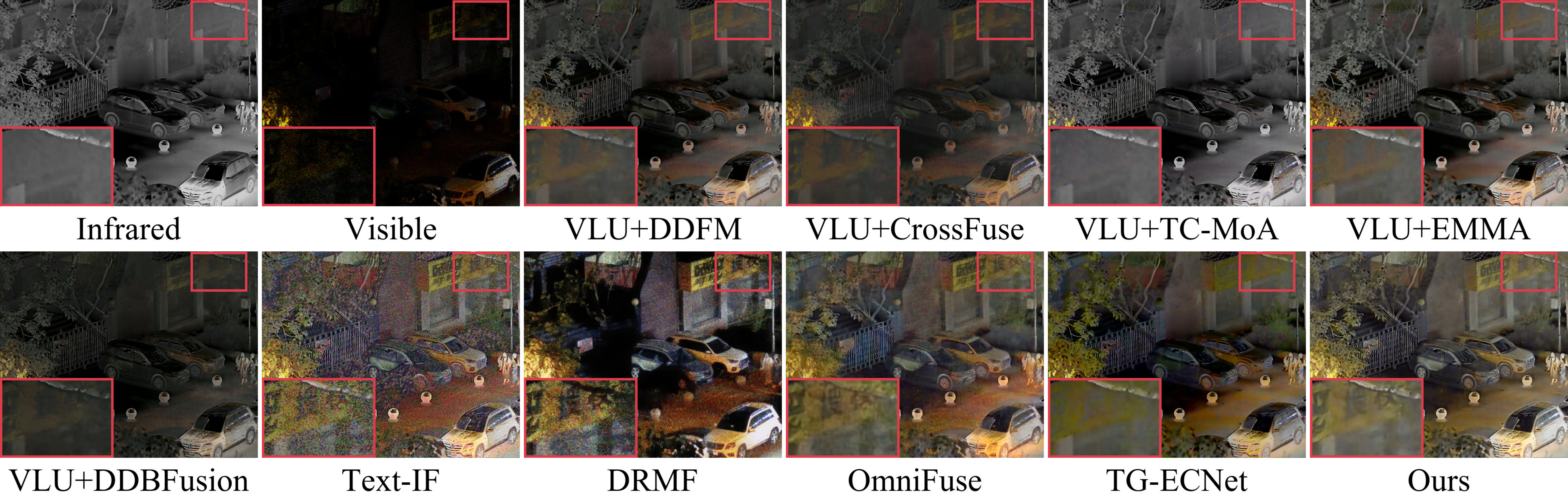}
  \caption{\rev{Results under compound degradation on LLVIP, where low light and defocus blur are applied to the visible image simultaneously. The insets enlarge the red-boxed regions.}}
  \label{fig:compound_qual}
\end{figure}

\rev{This setting uses 100 image pairs from LLVIP, where low-light corruption and defocus blur are applied simultaneously to the visible image while the infrared image stays intact. This combination never appears during training, so the test probes whether the framework handles two coupled corruptions rather than a single degradation family; results are given in Fig.~\ref{fig:compound_qual} and Tab.~\ref{tab:gen}(b) of the main text.}

\rev{Fig.~\ref{fig:compound_qual} shows that the baselines over-brighten the scene while leaving residual blur, or sharpen edges at the cost of amplified noise and color distortion; MODULE instead recovers the lost sharpness from the intact infrared structure without amplifying the compounded artifacts. Tab.~\ref{tab:gen}(b) shows the same trend: MODULE attains the best structural similarity, information transfer and gradient fidelity and remains among the top performers in the remaining metrics, while the baselines lose more accuracy as the corruption departs from the training distribution. The advantage in SSIM indicates that removing luminance and sharpness cues together does not disrupt the cross-modal structural prior exploited by the fusion branch.}

\section{Additional Ablation Studies}
\label{app:ablations}
\rev{This section reports three additional ablations that the main text omits for space: the initialization strategy of the iterative process (Tab.~\ref{tab:init}), the fidelity constraint $L_u$ (Tab.~\ref{tab:lu}), and the training behaviour with respect to the unfolding depth (Tab.~\ref{tab:kloss}).}

%\begin{table*}[!h]
%  \centering
%  \small
%  \renewcommand{\arraystretch}{1.0}
%  \caption{\rev{Effect of the initialization strategy of the iterative process} ({\color{red}best}, {\color{blue}second-best}).}
%  \label{tab:init}
%  \begin{tabular}{l@{\,~~~~}|l@{\,~~~}|c@{\,~~~}c@{\,~~~}c@{\,~~~}c}
%    \toprule
%    \rev{Setting} & \rev{Initialization} & \rev{SCD $\uparrow$} & \rev{VIF $\uparrow$} & \rev{$Q^{abf}$ $\uparrow$} & \rev{SSIM $\uparrow$} \\
%    \midrule
%    \multirow{3}{*}{IR Gaussian} & Mean & \textcolor{blue}{1.449} & 0.400 & 0.407 & 0.731 \\
%      & Conv & 1.397 & \textcolor{blue}{0.411} & \textcolor{blue}{0.422} & \textcolor{blue}{0.765} \\
%      & \textbf{Max (Ours)} & \textcolor{red}{1.473} & \textcolor{red}{0.447} & \textcolor{red}{0.433} & \textcolor{red}{0.771} \\
%    \midrule
%    \multirow{3}{*}{VIS low light} & Mean & 1.405 & \textcolor{blue}{0.572} & \textcolor{blue}{0.475} & \textcolor{blue}{0.789} \\
%      & Conv & \textcolor{blue}{1.409} & 0.563 & 0.462 & 0.780 \\
%      & \textbf{Max (Ours)} & \textcolor{red}{1.416} & \textcolor{red}{0.576} & \textcolor{red}{0.490} & \textcolor{red}{0.793} \\
%    \bottomrule
%  \end{tabular}
%\end{table*}

\rev{\textbf{\textit{i) Initialization Strategy}.}} \rev{Tab.~\ref{tab:init} shows that replacing the pixel-wise maximum selection by averaging or a learnable convolution does not improve the final fusion, and the maximum rule is consistently the best. This is expected: only a structural initialization is required for the first stage, and the element-wise maximum preserves the salient response of either modality without introducing additional parameters.}

%\begin{table*}[!h]
%  \centering
%  \small
%  \renewcommand{\arraystretch}{1.0}
%  \caption{\rev{Effect of the fidelity constraint $L_u$ on fusion and restoration} ({\color{red}best}).}
%  \label{tab:lu}
%  \begin{tabular}{l@{\,~~~~}|l@{\,~~~}|c@{\,~~~}c@{\,~~~}c@{\,~~~}c@{\,~~~}c@{\,~~~}c}
%    \toprule
%    \rev{Setting} & \rev{Variant} & \rev{SCD $\uparrow$} & \rev{VIF $\uparrow$} & \rev{$Q^{abf}$ $\uparrow$} & \rev{PSNR $\uparrow$} & \rev{SSIM $\uparrow$} & \rev{LPIPS $\downarrow$} \\
%    \midrule
%    \multirow{2}{*}{IR Gaussian} & w/o $L_u$ & \textcolor{red}{1.476} & 0.440 & 0.432 & 36.324 & 0.943 & 0.088 \\
%      & Ours & 1.473 & \textcolor{red}{0.447} & \textcolor{red}{0.433} & \textcolor{red}{36.422} & \textcolor{red}{0.944} & \textcolor{red}{0.087} \\
%    \midrule
%    \multirow{2}{*}{VIS low light} & w/o $L_u$ & 1.406 & 0.567 & 0.470 & 27.123 & 0.883 & 0.160 \\
%      & Ours & \textcolor{red}{1.416} & \textcolor{red}{0.576} & \textcolor{red}{0.490} & \textcolor{red}{27.377} & \textcolor{red}{0.891} & \textcolor{red}{0.138} \\
%    \bottomrule
%  \end{tabular}
%\end{table*}

\rev{\textbf{\textit{ii) Fidelity Constraint $L_u$}.}} \rev{Tab.~\ref{tab:lu} shows that the fidelity constraint mainly benefits the visible branch, which is the more severely degraded one: removing it degrades both the fused output and the restored visible image, and the perceptual distance grows markedly. On the infrared branch the effect is marginal and the difference in one metric falls within the rounding level, which is consistent with the observation that an already mild degradation leaves little room for the constraint to act.}

\begin{table*}[!h]
\centering
\renewcommand{\arraystretch}{1.0}

% ================= Left Table =================
\begin{minipage}[h]{0.45\textwidth}
  \centering
  \vspace{0pt}
  \captionof{table}{\rev{Effect of the initialization strategy of the iterative process} ({\color{red}best}, {\color{blue}second-best}).}
  \vspace{-0.05in}
  \label{tab:init}
  \resizebox{\linewidth}{!}{%
  \renewcommand{\arraystretch}{1.1}
    \begin{tabular}{l@{\,~}|l@{\,~}|c@{\,~~}c@{\,~~}c@{\,~~}c}
      \toprule
      \rev{Setting} & \rev{Initialization}
      & \rev{SCD $\uparrow$}
      & \rev{VIF $\uparrow$}
      & \rev{$Q^{abf}$ $\uparrow$}
      & \rev{SSIM $\uparrow$} \\
      \midrule
      \multirow{3}{*}{IR Gaussian}
        & Mean & \textcolor{blue}{1.449} & 0.400 & 0.407 & 0.731 \\
        & Conv & 1.397 & \textcolor{blue}{0.411} & \textcolor{blue}{0.422} & \textcolor{blue}{0.765} \\
        & \textbf{Max (Ours)} & \textcolor{red}{1.473} & \textcolor{red}{0.447} & \textcolor{red}{0.433} & \textcolor{red}{0.771} \\
      \midrule
      \multirow{3}{*}{VIS low light}
        & Mean & 1.405 & \textcolor{blue}{0.572} & \textcolor{blue}{0.475} & \textcolor{blue}{0.789} \\
        & Conv & \textcolor{blue}{1.409} & 0.563 & 0.462 & 0.780 \\
        & \textbf{Max (Ours)} & \textcolor{red}{1.416} & \textcolor{red}{0.576} & \textcolor{red}{0.490} & \textcolor{red}{0.793} \\
      \bottomrule
    \end{tabular}%
  }
\end{minipage}
\hfill
% ================= Right Table =================
\begin{minipage}[h]{0.53\textwidth}
  \centering
  \vspace{0pt}
  \captionof{table}{\rev{Effect of the fidelity constraint $L_u$ on fusion and restoration} ({\color{red}best}).}
  \vspace{-0.05in}
  \label{tab:lu}
  \resizebox{\linewidth}{!}{
  \renewcommand{\arraystretch}{1.3}
    \begin{tabular}{l@{\,~}|l@{\,~}|c@{\,~~}c@{\,~~}c@{\,~~}c@{\,~~}c@{\,~~}c}
      \toprule
      \rev{Setting} & \rev{Variant}
      & \rev{SCD $\uparrow$}
      & \rev{VIF $\uparrow$}
      & \rev{$Q^{abf}$ $\uparrow$}
      & \rev{PSNR $\uparrow$}
      & \rev{SSIM $\uparrow$}
      & \rev{LPIPS $\downarrow$} \\
      \midrule
      \multirow{2}{*}{IR Gaussian}
        & w/o $L_u$ & \textcolor{red}{1.476} & 0.440 & 0.432 & 36.324 & 0.943 & 0.088 \\
        & Ours & 1.473 & \textcolor{red}{0.447} & \textcolor{red}{0.433} & \textcolor{red}{36.422} & \textcolor{red}{0.944} & \textcolor{red}{0.087} \\
      \midrule
      \multirow{2}{*}{VIS low light}
        & w/o $L_u$ & 1.406 & 0.567 & 0.470 & 27.123 & 0.883 & 0.160 \\
        & Ours & \textcolor{red}{1.416} & \textcolor{red}{0.576} & \textcolor{red}{0.490} & \textcolor{red}{27.377} & \textcolor{red}{0.891} & \textcolor{red}{0.138} \\
      \bottomrule
    \end{tabular}%
  }
\end{minipage}

\end{table*}

\rev{\textbf{\textit{iii) Unfolding Depth and Training Losses}.}} \rev{Deeper unfolding keeps reducing the training losses in Tab.~\ref{tab:kloss} ($L_{fuse}$ and the projection, fidelity and restoration terms of both branches), yet the test performance in Tab.~\ref{tab:ablation_by_degradation} peaks at three stages and slightly declines afterwards. This gap between training and test behaviour indicates that the additional stage starts to fit the synthetic training degradations rather than the underlying degradation model, which is why we adopt three stages as a compromise between representation capacity and generalization.}
\begin{table*}[h]
\centering

% ================= Left Table =================
\begin{minipage}[h]{0.45\textwidth}
  \centering
  \vspace{0pt}
  \small
  \renewcommand{\arraystretch}{1.2}
  \captionof{table}{\rev{Training losses of the final stage under different unfolding depths} ({\color{red}lowest}).}
  \label{tab:kloss}
  \vspace{-0.05in}
  \resizebox{\linewidth}{!}{%
    \begin{tabular}{c@{\,~~~}|c@{\,~~~}c@{\,~~~}c@{\,~~~}c@{\,~~~}c@{\,~~~}c@{\,~~~}c}
      \toprule
      \rev{$K$} & \rev{$L_{fuse}$} & \rev{$L_{pi}$} & \rev{$L_{pv}$} & \rev{$L_{ui}$} & \rev{$L_{uv}$} & \rev{$L_{zi}$} & \rev{$L_{zv}$} \\
      \midrule
      3 & 2.626 & 0.011 & 0.009 & \textcolor{red}{0.005} & 0.025 & \textcolor{red}{0.014} & 0.016 \\
      4 & \textcolor{red}{2.586} & \textcolor{red}{0.007} & \textcolor{red}{0.006} & 0.033 & \textcolor{red}{0.006} & 0.016 & \textcolor{red}{0.014} \\
      \bottomrule
    \end{tabular}%
  }
\end{minipage}
\hfill
% ================= Right Table =================
\begin{minipage}[h]{0.53\textwidth}
  \centering
  \vspace{0pt}
  \small
  \renewcommand{\arraystretch}{1.1}
  \captionof{table}{\rev{Restoration quality under feedback removal}.}
  \label{tab:restoration_branch}
  \vspace{-0.05in}
  \resizebox{\linewidth}{!}{%
    \begin{tabular}{l@{\,~}|c@{\,~~}c@{\,~~}c@{\,~}|c@{\,~~}c@{\,~~}c@{\,~~}}
      \toprule
      \multirow{2}{*}[-0.5ex]{\textbf{Variant}}
      & \multicolumn{3}{c|}{\textbf{Visible}}
      & \multicolumn{3}{c}{\textbf{Infrared}} \\
      \cmidrule(lr){2-4} \cmidrule(lr){5-7}
      & PSNR $\uparrow$ & SSIM $\uparrow$ & LPIPS $\downarrow$
      & PSNR $\uparrow$ & SSIM $\uparrow$ & LPIPS $\downarrow$ \\
      \midrule
      w/o $G_m$
      & {\color{blue}15.674} & {\color{blue}0.817} & {\color{blue}0.196}
      & 36.388 & 0.943 & 0.088 \\
      w/o $L_p$
      & 13.545 & 0.790 & 0.233
      & {\color{blue}36.401} & {\color{blue}0.943} & {\color{red}0.082} \\
      \textbf{Ours}
      & {\color{red}27.377} & {\color{red}0.891} & {\color{red}0.138}
      & {\color{red}36.422} & {\color{red}0.944} & {\color{blue}0.087} \\
      \bottomrule
    \end{tabular}%
  }
\end{minipage}
\end{table*}

%\begin{table*}[!h]
%  \centering
%  \small
%  \renewcommand{\arraystretch}{1.0}
%  \caption{\rev{Training losses of the final stage under different unfolding depths} ({\color{red}lowest}).}
%  \vspace{-0.05in}
%  \label{tab:kloss}
%  \resizebox{0.6\linewidth}{!}{
%  \begin{tabular}{c@{\,~~~}|c@{\,~~~}c@{\,~~~}c@{\,~~~}c@{\,~~~}c@{\,~~~}c@{\,~~~}c}
%    \toprule
%    \rev{$K$} & \rev{$L_{fuse}$} & \rev{$L_{pi}$} & \rev{$L_{pv}$} & \rev{$L_{ui}$} & \rev{$L_{uv}$} & \rev{$L_{zi}$} & \rev{$L_{zv}$} \\
%    \midrule
%    3 & 2.626 & 0.011 & 0.009 & \textcolor{red}{0.005} & 0.025 & \textcolor{red}{0.014} & 0.016 \\
%    4 & \textcolor{red}{2.586} & \textcolor{red}{0.007} & \textcolor{red}{0.006} & 0.033 & \textcolor{red}{0.006} & 0.016 & \textcolor{red}{0.014} \\
%    \bottomrule
%  \end{tabular}}
%\end{table*}

\rev{\textbf{\textit{iv) Restoration under Feedback Removal}.}} \rev{Tab.~\ref{tab:restoration_branch} reports the restoration quality of the intermediate estimates when the fusion-to-restoration feedback is disabled. Disabling the feedback markedly degrades the visible restoration, and removing only the projection consistency loss degrades it further, since an unconstrained projection writes erroneous structure back. The infrared branch changes only marginally across the variants, which is consistent with its much milder degradation.}
%
%\begin{table*}[!h]
%  \centering
%  \caption{\rev{Restoration quality under feedback removal} ({\color{red}best}, {\color{blue}second-best}).}
%  \vspace{-0.05in}
%  \label{tab:restoration_branch}
%  \resizebox{0.7\linewidth}{!}{
%    \renewcommand{\arraystretch}{1.1}
%    \begin{tabular}{l@{\,~~~~}|c@{\,~~~~}c@{\,~~~~}c@{\,~~~~}|c@{\,~~~~}c@{\,~~~~}c@{\,~~~}}
%    \toprule
%    \multirow{2}{*}[-0.5ex]{\textbf{Variant}} & \multicolumn{3}{c|}{\textbf{Visible}} & \multicolumn{3}{c}{\textbf{Infrared}} \\
%    \cmidrule(lr){2-4} \cmidrule(lr){5-7}
%    & PSNR $\uparrow$ & SSIM $\uparrow$ & LPIPS $\downarrow$ & PSNR $\uparrow$ & SSIM $\uparrow$ & LPIPS $\downarrow$ \\
%    \midrule
%    w/o $G_m$          & {\color{blue}15.674} & {\color{blue}0.817} & {\color{blue}0.196} & 36.388 & 0.943 & 0.088 \\
%    w/o $L_p$          & 13.545 & 0.790 & 0.233 & {\color{blue}36.401} & {\color{blue}0.943} & {\color{red}0.082} \\
%    \textbf{Ours}      & {\color{red}27.377} & {\color{red}0.891} & {\color{red}0.138} & {\color{red}36.422} & {\color{red}0.944} & {\color{blue}0.087} \\
%    \bottomrule
%    \end{tabular}%
%  }
%\end{table*}

\section{Detailed Degradation Generation}
\label{app:degradation}
\rev{This section describes how the synthetic degradations used throughout the paper are generated, including the formulation of each degradation type and the range of its parameters. All degraded--clean pairs are synthesized from clean LLVIP and MSRS images, and each pair is degraded exactly once, with every parameter independently sampled from a fixed distribution.}

\rev{\textbf{Gaussian noise (both modalities).} $I_{noisy}=I+\mathcal{N}(0,\sigma^2)$ with $\sigma\sim U(5,30)$. Noise is added to each RGB channel independently for visible image and to the single channel for infrared image.}

\rev{\textbf{Stripe noise (infrared).} $I_{noisy}(x,y)=I(x,y)+S(y)$ with $S(y)\sim\mathcal{N}(0,\beta^2)$ and $\beta\sim U(3,8)$; a single one-dimensional noise vector is shared by all rows, which reproduces the column-wise pattern of readout electronics.}

\rev{\textbf{Defocus blur (visible).} $I_{blur}=I\ast K$, where $K$ is a normalized circular point-spread function of size $k\times k$ with radius $r=\lfloor k/2\rfloor$ and $k\in\{5,7,9,11\}$.}

\rev{\textbf{Low light with shot noise (visible).} The illumination is first attenuated by $\tilde{I}=(I/255)^{\gamma}$ with $\gamma\sim U(2.0,3.5)$, and sensor noise is then injected following a Poisson model $I_{noisy}=(255/P)\cdot\mathcal{P}(\tilde{I}\cdot P)$ with $P\sim U(15,35)$; a smaller $P$ corresponds to a lower signal-to-noise ratio.}

\rev{Training and testing pairs are drawn independently from these distributions. The compound setting in Appendix~\ref{app:compound} combines two visible corruptions that never co-occur during training.}

\section{Medical Evaluation}
\begin{figure}[!h]
     \centering
        \includegraphics[width=\linewidth]{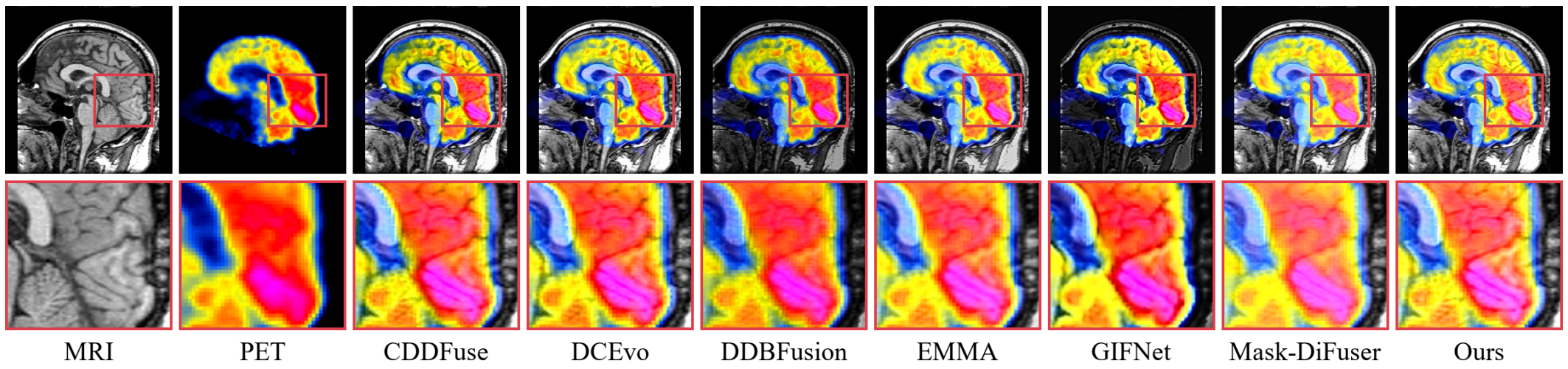}\\
        \vspace{-0.05in}
     \caption{Qualitative results in MRI-PET image fusion.}
     \label{fig:medical}
\end{figure}

 We select 100 MRI-PET image pairs from the Harvard Medical dataset\footnote{http://www.med.harvard.edu/AANLIB/home.html} for training and 50 pairs for testing. For competitors, we select methods capable of performing medical image fusion. As Fig.~\ref{fig:medical} shows, our method exhibits a superior capability in structural and texture restoration, preserving sharp anatomical edges and fine-grained diagnostic details that are often blurred by other SOTA methods.

As Tab.~\ref{tab:medical_results} shows, our method achieves advantages across all evaluated metrics. The results underscore the benefit of our generative modeling approach. By bypassing the limitations of hand-crafted modal-specific priors, our framework robustly handles the distinct imaging physics of medical modalities, providing a unified and high-performance solution for multi-modal image fusion.
\begin{table*}[!h]
  \centering
  \renewcommand{\arraystretch}{1.0}
  \caption{Quantitative comparison on medical image fusion ({\color{red}best}, {\color{blue}second-best}).}
  \vspace{-0.05in}
  \label{tab:medical_results}
  \resizebox{\linewidth}{!}{%
  \begin{tabular}{l@{\,~~~~}|c@{\,~~~~}c@{\,~~~~}c@{\,~~~~}c@{\,~~~~}c@{\,~~~~}c@{\,~~~~}c@{\,~~~~}}
    \toprule
    Metrics & CDDFuse~\cite{cdd} & EMMA~\cite{emma} & GIFNet~\cite{gif} & DCEvo~\cite{dcevo} & DDBFusion~\cite{DDBFusion} & Mask-Difuser~\cite{mask} & \textbf{Ours} \\
    \midrule
    SCD $\uparrow$        & 1.529 & 1.471 & 1.296 & 1.478 & 1.271 & \textcolor{red}{1.619} & \textcolor{blue}{1.616} \\
    VIF $\uparrow$        & 0.776 & 0.565 & 0.416 & \textcolor{blue}{0.787} & 0.561 & 0.759 & \textcolor{red}{0.793} \\
    $Q_{abf}$ $\uparrow$ & \textcolor{blue}{0.751} & 0.610 & 0.339 & 0.742 & 0.508 & 0.694 & \textcolor{red}{0.753} \\
    SSIM $\uparrow$       & \textcolor{blue}{0.632} & 0.370 & 0.225 & 0.373 & 0.503 & 0.373 & \textcolor{red}{0.726} \\
    \bottomrule
  \end{tabular}%
  }
\end{table*}

\section{Coefficient Analysis}
As the core contribution is the restoration-fusion mutual-promoting mechanism, we analyze coefficients of losses critical to this mechanism.

For fusion loss, the gradient loss preserves perceptually important structures and high-frequency textures, making $\alpha_2$ highly influential for fusion performance. Moreover, via $G_m$, these structures are back-projected to guide restoration, making $\alpha_2$ also critical for the joint system. Other coefficients are fixed to keep the output usable. For consistency loss, $\lambda_p$ of $L_p$ controls our novel inter-task mutual promotion mechanism. Analyzing it reveals the effect of this mechanism on performance and helps find a balanced regime. Other consistency losses are tied to the physical meanings of intermediate variables and thus fixed to avoid confounding.

\begin{figure}[!h]
     \centering
        \includegraphics[width=\linewidth]{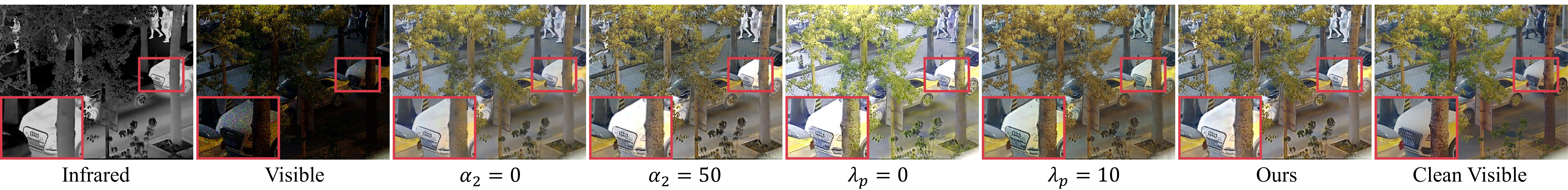}\\
        \vspace{-0.05in}
     \caption{Qualitative results of coefficient analysis.}
     \label{fig:coefficient}
\end{figure}

As Fig.~\ref{fig:coefficient} shows,  \(\alpha_2\) balances sharpness and fidelity: \(\alpha_2=0\) loses high-frequency details, while \(\alpha_2=50\) over-amplifies gradients, causing color distortion. The coupling term \(\lambda_p\): \(\lambda_p=0\) decouples tasks, leading to color shift; \(\lambda_p=10\) over-restricts flexibility. The framework remains stable even at extreme values due to the inductive bias and mutual regularization of deep unfolding.

Quantitatively, as Tab.~\ref{tab:coefficient} shows, increasing \(\alpha_2\) improves SCD and \(Q^{abf}\) but reduces SSIM and VIF. Removing the coupling term (\(\lambda_p=0\)) causes a performance collapse across all metrics. The dynamic strategy \(\lambda_p = 2 + 0.1k\) outperforms a fixed large coefficient. The configuration \(\alpha_2 = 8\) with dynamic \(\lambda_p\) achieves the best balance, yielding optimal overall performance.
\begin{table*}[!h]
  \centering
  \renewcommand{\arraystretch}{1.0}
  \caption{Ablation study on hyper-parameters $\alpha_2$ and $\lambda_p$ ($k$: $k$-th iteration) ({\color{red}best}, {\color{blue}second-best}).}

  \label{tab:coefficient}
  \resizebox{0.65\linewidth}{!}{%
  \begin{tabular}{l@{\,~~~~~~~}|c@{\,~~~~~~~}c@{\,~~~~~~~}c@{\,~~~~~~~}c@{\,~~~}}
    \toprule
    Coefficients & SCD $\uparrow$ & VIF $\uparrow$ & Qabf $\uparrow$ & SSIM $\uparrow$ \\
    \midrule
    $\alpha_2 = 0, \lambda_p = 2 + 0.1k$ & 1.350 & 0.572 & 0.404 & 0.836 \\
    $\alpha_2 = 50, \lambda_p = 2 + 0.1k$ & \textcolor{red}{1.699} & \textcolor{blue}{0.584} & \textcolor{red}{0.512} & \textcolor{blue}{0.847} \\
    $\alpha_2 = 8, \lambda_p = 10$ & 1.349 & 0.583 & 0.462 & 0.811 \\
    $\alpha_2 = 8, \lambda_p = 0$ & 1.284 & 0.384 & 0.368 & 0.631 \\
    $\alpha_2 = 8, \lambda_p = 2 + 0.1k$ & \textcolor{blue}{1.516} & \textcolor{red}{0.591} & \textcolor{blue}{0.498} & \textcolor{red}{0.873} \\
    \bottomrule
  \end{tabular}}
\end{table*}

\section{Theoretical Analysis of the Surrogate Objective}
\label{app:theory}
We provide a theoretical analysis of the surrogate objective used in our deep unfolding framework. \rev{The following analysis concerns the exact pseudo-inverse that defines the surrogate objective; the learned realization inherits the resulting conditioning and the explicit separation of the null-space component, not the algebraic identities of $D_m^\dagger$.}
In deep unfolding networks (DUNs), the fidelity objective essentially shapes the optimization landscape, which inherently guides the structural prior and the solution preference of the network. The conventional objective function $\frac{1}{2} \|y_m - D_m z_m\|_F^2$ faces a specific limitation: it does not explicitly constrain the null-space $\mathcal{N}(D_m)$ of the degradation operator. Consequently, its optimization landscape tends to be flat along these null-space directions, lacking a clear preference for recovering unobservable degradation components. This can occasionally lead the iterative solver toward ambiguous solutions.
By introducing the surrogate objective $\frac{1}{2} \|D_m^\dagger y_m - z_m\|_F^2$, we can effectively reshape this landscape. We decompose the term inside the surrogate objective as:
\begin{equation}
D_m^\dagger y_m - z_m = D_m^\dagger(y_m - D_m z_m) - (I - D_m^\dagger D_m)z_m.
\end{equation}
We denote $\mathbf{a} = D_m^\dagger(y_m - D_m z_m)$ and $\mathbf{b} = (I - D_m^\dagger D_m)z_m$. According to the properties of the exact pseudoinverse, $\mathbf{a}$ lies in the row space $\mathcal{R}(D_m^T)$ of the degradation operator $D_m$, and $\mathbf{b}$ lies in the null space $\mathcal{N}(D_m)$ since $D_m(I - D_m^\dagger D_m) = 0$. Because the row space and null space are orthogonal complements ($\mathcal{R}(D_m^T) \perp \mathcal{N}(D_m)$), $\mathbf{a}$ and $\mathbf{b}$ are strictly orthogonal. Applying the Pythagorean theorem for the Frobenius norm yields:
\begin{equation}
\frac{1}{2} \|D_m^\dagger y_m - z_m\|_F^2 = \frac{1}{2} \|D_m^\dagger(y_m - D_m z_m)\|_F^2 + \frac{1}{2} \|(I - D_m^\dagger D_m) z_m\|_F^2.
\end{equation}
This decomposition reveals the dual mechanism of the surrogate objective.
The first right term, $\frac{1}{2} \|D_m^\dagger(y_m - D_m z_m)\|_F^2$, mitigates the impact of small singular values of $D_m$, smoothing the narrow valleys into a more well-conditioned landscape.
The second term, $\frac{1}{2} \|(I - D_m^\dagger D_m) z_m\|_F^2$, explicitly imposes a null-space regularization.
This reshaped landscape induces a mathematical preference for stable, valid solutions, providing a more stable geometric prior for the subsequent task decomposition and encouraging a robust restoration trajectory.
Motivated by these profound mathematical properties, our unfolding framework explicitly operates on this surrogate objective, leveraging its robust trajectory to guide the multi-stage iterative solver.

\section{Limitations}
\label{sec:limitations}
This work handles degradations from both the infrared and visible modality. While the proposed framework effectively addresses common degradations, its robustness might decrease when confronting highly complex or completely unknown hybrid degradations in unpredictable open-world scenarios. Future work will explore restoration techniques for more degradations to extend its applicability to a broader spectrum of extreme degradations.

\end{document}